\documentclass[10pt]{article} % For LaTeX2e
\usepackage[preprint]{tmlr}

\usepackage{amsmath,amsfonts,bm}

\def\eqref#1{equation~\ref{#1}}
\def\1{\bm{1}}

\def\mI{{\bm{I}}}

\DeclareMathAlphabet{\mathsfit}{\encodingdefault}{\sfdefault}{m}{sl}
\SetMathAlphabet{\mathsfit}{bold}{\encodingdefault}{\sfdefault}{bx}{n}

\newcommand{\E}{\mathbb{E}}

\newcommand{\R}{\mathbb{R}}

\newcommand{\Var}{\mathrm{Var}}

\newcommand{\Cov}{\mathrm{Cov}}
\DeclareMathOperator*{\argmin}{arg\,min}

\usepackage{hyperref}
\usepackage{url}

\usepackage[utf8]{inputenc} % allow utf-8 input
\usepackage[T1]{fontenc}    % use 8-bit T1 fonts
\usepackage{hyperref}       % hyperlinks
\usepackage{url}            % simple URL typesetting
\usepackage{booktabs}       % professional-quality tables
\usepackage{amsfonts}       % blackboard math symbols
\usepackage{nicefrac}       % compact symbols for 1/2, etc.
\usepackage{microtype}      % microtypography
\usepackage{xcolor}         % colors
\usepackage{amssymb, amsthm, amsmath}
\usepackage{centernot}
\usepackage[createShortEnv]{proof-at-the-end}
\usepackage{subcaption}
\usepackage{enumitem}
\theoremstyle{definition}
\newtheorem{definition}{Definition}[section]

\usepackage{graphicx} 

\usepackage[noabbrev]{cleveref}

\newcommand{\norm}[1]{\left\lVert#1\right\rVert}
\title{Beyond Pairwise Correlations: Higher-Order Redundancies in Self-Supervised Representation Learning}

\author{\name David Zollikofer$^*$ \email zdavid@ethz.ch \\
      \AND
      \name Béni Egressy$^*$ \email begressy@ethz.ch \\
      \addr Distributed Computing Group, ETH Zürich, Zürich, Switzerland
      \AND
      \name Frederik Benzing \email  benzingfr@gmail.com \\
      \AND
      \name Matthias Otth \email maotth@ethz.ch \\
      \AND
      \name Roger Wattenhofer \email wattenhofer@ethz.ch \\
      \addr Distributed Computing Group, ETH Zürich, Zürich, Switzerland
}

\def\month{MM}  % Insert correct month for camera-ready version
\def\year{YYYY} % Insert correct year for camera-ready version
\def\openreview{\url{https://openreview.net/forum?id=XXXX}} % Insert correct link to OpenReview for camera-ready version

\begin{document}

\maketitle
\def\thefootnote{*}\footnotetext{These authors contributed equally to this work.}\def\thefootnote{\arabic{footnote}}
\begin{abstract}
    Several self-supervised learning (SSL) approaches have shown that redundancy reduction in the feature embedding space is an effective tool for representation learning. However, these methods consider a narrow notion of redundancy, focusing on pairwise correlations between features. To address this limitation, we formalize the notion of embedding space redundancy and introduce redundancy measures that capture more complex, higher-order dependencies. We mathematically analyze the relationships between these metrics, and empirically measure these redundancies in the embedding spaces of common SSL methods. Based on our findings, we propose Self Supervised Learning with Predictability Minimization (SSLPM) as a method for reducing redundancy in the embedding space. SSLPM combines an encoder network with a predictor engaging in a competitive game of reducing and exploiting dependencies respectively. We demonstrate that SSLPM is competitive with state-of-the-art methods and find that 
    % all tested SSL methods outperforming SSLPM feature strictly less 
    the best performing SSL methods exhibit low
    embedding space redundancy, suggesting that even methods without explicit redundancy reduction mechanisms perform redundancy reduction implicitly. 
\end{abstract}

\section{Introduction}
\label{sec:introduction}

Self-supervised learning has been shown to produce image embeddings of similar quality as its supervised learning counterparts \citep{pmlr-v119-chen20j}. Early methods, such as SimCLR, achieved this by using positive and negative sample pairs within the loss function. A positive pair refers to two different (random) perturbations of the same input image, whereas a negative sample pair refers to perturbed versions of two different input images. The SimCLR loss function encourages similar representations for positive pairs and dissimilar ones for negative pairs. 
In contrast, subsequent work by \citet{zbontar2021barlow} replaced negative sample pairs in the loss function with covariance reduction in the representation space. Instead of using negative sample pairs, their loss function reduces the redundancy between representation dimensions by reducing pairwise correlations. 
% Somewhat surprisingly, this achieved similarly impressive results.

While pairwise decorrelating the representation space produces impressive results \citep{zbontar2021barlow}, it ignores potential redundancies that might remain. More precisely, the loss function does not explicitly reduce higher-order (involving more than two representation features) and non-linear redundancies in the representation space. 
% In light of recent theoretical work on compression in SSL \citep{e26030252} looking at redundancy between augmented views as well as their representations, 
In light of this observation, 
several natural questions arise: 
\begin{itemize}
    \item How can redundancy in SSL embedding spaces be quantified? %much redundancy can be found in SSL embedding spaces?
    \item How do redundancies in the embedding space affect downstream performance?
    \item Can performance be further improved by removing additional, more complex redundancies?
\end{itemize}
% (1) Which forms of redundancy can be found in embedding spaces? (2) How do redundancies in the embedding space affect downstream performance? (3) Can performance be further improved by removing additional, more complex redundancies?

To answer these research questions we: (1) Introduce formal definitions of embedding space redundancies, specifically pairwise, linear, and non-linear redundancies, and derive their theoretical relationships; (2) Propose Self-Supervised Learning with Predictability Minimization (SSLPM), a novel SSL method built on redundancy reduction, which we show to be competitive with current state of the art; (3) Empirically investigate the relationship between model performance and embedding space redundancy in a wide range of SSL methods including Barlow Twins \citep{zbontar2021barlow}, BYOL \citep{grill2020bootstrap}, NNCLR \citep{dwibedi2021little}, SimCLR \citep{pmlr-v119-chen20j}, MocoV3 \citep{he2020momentum,chen2021mocov3}, VICReg \citep{bardes2021vicreg}, and VIbCReg \citep{lee2021computer}.

In our experiments, we find that 
\begin{itemize}
    \item Reducing additional redundancies in training does not result in higher downstream performance.  
\item Models with explicit redundancy reduction, such as Barlow Twins or SSLPM, show a clear link between performance and linear redundancy. However, this link does not hold in general. 
\item All methods outperforming our SSLPM model exhibit strictly less embedding space redundancy, suggesting that even methods without explicit redundancy reduction perform redundancy reduction implicitly. However, for SSL methods in general, performance is only weakly correlated with embedding space redundancy.
\item The projector \footnote{Typically SSL networks use a projector between the embedding space used for downstream tasks and the representation space where the loss is calculated. The projector network consists of a stack of Feed-forward layers. Please refer to Figure \ref{fig:ssl_schematic}.} depth has a significant impact on redundancy reduction: more projector layers result in less linear and nonlinear redundancy in the embedding space. 
\end{itemize}

% In conclusion, although we demonstrate that high performing SSL methods feature lower redundancy, reducing redundancy further, and in particular reducing more complex forms of redundancy, does not improve performance. 
% Our detailed analysis offers novel insights into the role of redundancy in self supervised learning. 

\section{Related Work}
Early methods for SSL, such as SimCLR \citep{pmlr-v119-chen20j} or InfoNCE \citep{oord2018representation}, use a contrastive loss function based on positive and negative input pairs. The loss function incentivizes augmentations from the same input (positive pair) to produce similar embeddings, while pushing embeddings from different inputs (negative pairs) further apart in the embedding space. Follow-up works on SSL, such as BYOL \citep{grill2020bootstrap} and SiamSiam \citep{chen2020exploring}, have shown that negative samples are not always necessary by using asymmetric designs in their Siamese networks.

A key challenge in SSL is the prevention of model collapse, whereby all embeddings converge to the same point so as to reduce the positive pair loss. Model collapse leads to no useful representations being learned. 
Instead of using negative pairs explicitly, Barlow Twins \citep{zbontar2021barlow}, VicReg \citep{bardes2021vicreg}, VibCReg \citep{lee2021computer} and W-MSE \citep{ermolov2021whitening} address model collapse by decorrelating features in the representation space of the model. 
By decorrelating the features in the representation space, redundancy is reduced. 

The idea of representation space redundancy reduction was first introduced by \citet{schmidhuber1996semilinear} and \citet{schraudolph1999processing} through their concept of predictability minimization. In predictability minimization, an encoder transforms inputs to an $n$-dimensional representation space, on which $n$ predictors try to predict, in a leave-one-out fashion, every single representation feature from the others. The encoder tries to minimize redundancy by ensuring predictors cannot achieve a low prediction loss. The idea of two networks engaging in a competitive game was later revisited in the context of Generative Adversarial Networks \citep{goodfellow2014generative, schmidhuber2020generative}. 

Nevertheless, predictors are not novel to SSL and have found applications in \citep{assran2023self} where they aim to reconstruct representations of surrounding image patches from a single patch.

Moreover, recent work by \citet{e26030252} has drawn connections between compression in SSL methods and information theory. Our work, empirical in nature offers new insights in the role of redundancy reduction in SSL.

\section{Methods}
\label{sec:methods}

\subsection{Background}

\begin{figure}
    \centering
    \includegraphics[width=0.8\linewidth]{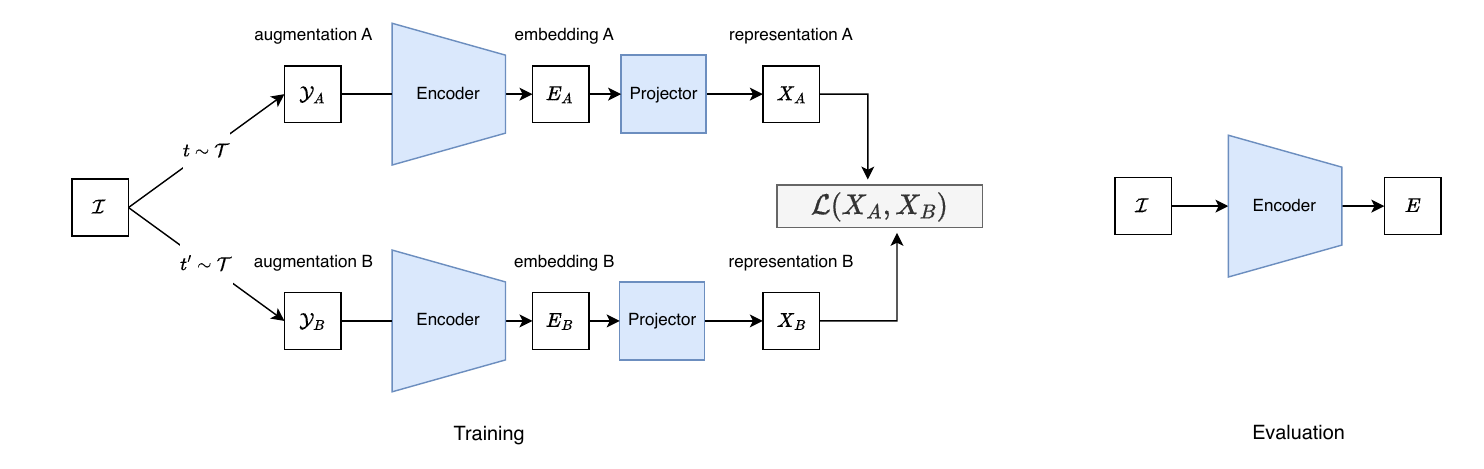}
    \caption{Siamese training set up with encoder in self supervised learning. Embeddings are used in downstream applications, whereas representations are used for loss calculation during pretraining. After pretraining, during evaluation, only the encoder is kept and the randomized augmentations $\tau$ are not applied.}
    \label{fig:ssl_schematic}
\end{figure}

Our work is inspired by and builds on Barlow Twins \citep{zbontar2021barlow}.
Barlow Twins is a self-supervised learning approach based on the Siamese training setup depicted in Figure \ref{fig:ssl_schematic}. It is used to pretrain neural networks before these are applied to downstream tasks. 

In a Siamese network, an input batch $\mathcal{I}$ is passed through two different transformations $\mathbf{t}$ and $\mathbf{t'}$ drawn at random from a distribution of transformations $\mathcal{T}$, producing augmentations $\mathcal{Y_A}$ and $\mathcal{Y_B}$. The augmentations are then passed though a neural network, consisting of an encoder and a projector in series, yielding two batches of representations $\mathbf{X_A}$ and $\mathbf{X_B}$. 

The Barlow Twins loss function, as displayed in Equation \ref{eq:barlowTwinsLoss}, uses these representations to compute the empirical covariance matrix $\mathbf{X_A} ^T \mathbf{X_B}$. Before doing so, each dimension of the representations is centered and standardized within its respective batch. 
The on-diagonal elements of the covariance matrix are then used for the invariance loss $\mathcal{L}_\text{invariance}$, whereby each representation dimension is optimized to have a correlation of 1 regardless of the augmentation. 
Meanwhile, the redundancy reduction term $\mathcal{L}_\text{redundancy reduction}$ minimizes the off-diagonal elements of the empirical covariance matrix, which effectively pairwise decorrelates the representation dimensions.
\begin{align}\label{eq:barlowTwinsLoss}
\mathcal{L}_{\text{Barlow Twins}} = \underbrace{\sum_i (1 - \left(\mathbf{X_A} ^T \cdot \mathbf{X_B}\right)_{ii})^2}_{\mathcal{L}_\text{invariance}} + \lambda \underbrace{\sum_i \sum_{j \neq i} \left(\mathbf{X_A} ^T \cdot \mathbf{X_B}\right)_{ij}^2}_{\mathcal{L}_\text{redundancy reduction}}
\end{align}

Similar to contrastive learning (CL) methods, Barlow Twins aims to pull the representations from the same input sample (positive pairs) closer together. In this way, the encoder learns to become invariant to the augmentations in $\mathcal{T}$. This is achieved by forcing the diagonal elements of the correlation matrix towards $1$. However, unlike CL methods which push representations from different input samples (negative pairs) further apart, Barlow Twins focuses on reducing the redundancy between the individual features of $\mathbf{X_A}$ and $\mathbf{X_B}$. This is achieved by forcing the off-diagonal of the correlation matrix towards $0$.

After the pretraining phase has been completed, the projector is removed and only the encoder is used to embed inputs for downstream tasks. A prediction head, often a single linear layer, is added to produce the desired output format. Fine-tuning is typically done only on the prediction head, and we will also follow this process in this paper.

\subsection{Redundancy Measures}

Inspired by the pairwise correlation redundancy reduction term, we introduce a set of redundancy measures with the aim of capturing additional redundancy between the representation features.

Let $\mathbf{X} = [\textnormal{X}_1,\ldots,\textnormal{X}_n]^T$ be a vector of $n$ centered random variables, with each variable $\textnormal{X}_i$ representing the value of the $i$-th dimension of the $n$ dimensional embedding space. We assume $\E[\textnormal{X}_i] = 0$ and $\Var[\textnormal{X}_i] = 1$ which is achieved by standardization over the dataset. For convenience we define  $\mathbf{X}_{-j} = [\textnormal{X}_1,\ldots,\textnormal{X}_{j-1} ,\textnormal{X}_{j+1} \ldots \textnormal{X}_n]^T$

\begin{definition}[Average Absolute Covariance]
    For the $n$ dimensional random vector $\mathbf{X}$ with each dimension satisfying $\E[\textnormal{X}_i] = 0$ and $\Var[\textnormal{X}_i] = 1$, we define the average absolute covariance of dimension $j$ ($\operatorname{AAC}_j$) as well as the total average absolute covariance ($\operatorname{AAC}$) as 
\begin{align}
\operatorname{AAC}_j(\mathbf{X}) &:= \frac{\sum_{i= 1 :  i\neq j}^n |\Cov(\textnormal{X}_i,\textnormal{X}_j)|}{(n-1)} & \quad    \operatorname{AAC}(\mathbf{X}) &:= \frac{1}{n} \sum_{j=1}^n \operatorname{AAC}_j(\mathbf{X})
\end{align}

\end{definition}

Inspired by the concept of predictability minimization by \citet{schmidhuber1996semilinear}, we complement the definition of AAC with a definition of predictability, where instead of measuring the pairwise correlation between variables, we measure to what degree a variable can be reconstructed from all others using a \emph{predictor} network. 

\begin{definition}[Predictability]
    For the $n$ dimensional random vector $\mathbf{X}$ with each dimension $\E[\textnormal{X}_i] = 0$, $\Var[\textnormal{X}_i] = 1$, we define the predictability of index $j$ ($\operatorname{Pred}_j$) and the total predictability ($\operatorname{Pred}$) as
    \begin{align}
     \operatorname{Pred}_j (\textnormal{X}_j|\mathbf{X}_{-j}) &:= 1-  \min_{F\in\mathcal{F}} \E_{\textnormal{X}_i, 1 \leq i \leq n} \left[ || F(\mathbf{X}_{-j}) - \textnormal{X}_{j} ||_2^2 \right] &\quad \operatorname{Pred}(\mathbf{X})&:= \frac{1}{n} \sum_{j=1}^n \operatorname{Pred}_j (\textnormal{X}_j|\mathbf{X}_{-j})
    \end{align}
where $\mathcal{F} := \{F : \R^{n-1} \to \R\}$ is the set of all possible functions.
\end{definition}
Given this definition, determining predictability is intractable in general since we minimize over the set of all functions $\mathcal{F}$. We therefore introduce two approximations:
% Linear Redundancy (LR) restricts the function space to linear functions and Non-Linear Redundancy (NLR) 

% Therefore, we introduce to approximations of it... For Linear Predictability we restirct the function space to be linear functions...and for Nnon-linear predictability we restrict to functions parametrized by neural networks; in our case 3 layer mlps with x hidden units."
\begin{itemize}
    \item {Linear Redundancy (LR)} restricts the function space $\mathcal{F}$ to linear functions: \begin{align}
    \operatorname{LR}_j (\mathbf{X}) &=1 - \min_{\vec{b}}  \E \left[|| \mathbf{X}_{-j} \cdot \vec{b} - \textnormal{X}_j ||_2 ^2  \right] &\quad  \operatorname{LR}(\mathbf{X}) &= \frac{1}{n} \sum_{j=1}^n \operatorname{LR}_j(\mathbf{X}) 
\end{align}
with $\vec{b}$ representing the coefficients of a linear regression. In practice, we deviate from this for numerical stability by using ridge regression with penalty $\mu$, with $\mu$ being chosen through cross-validation. 
    \item {{Non-linear Redundancy (NLR)}} restricts the function space to neural networks $\operatorname{MLP}_\theta$ parameterized by weights and biases $\theta$.
    \begin{align}  
    \operatorname{NLR}_j(\mathbf{X}) &= 1 - \min_\theta \E \left[  || \operatorname{MLP}_\theta\left(\mathbf{X}_{-j} \right) - \textnormal{X}_j ||_2 ^2 \right] &\quad \operatorname{NLR}(\mathbf{X}) &= \frac{1}{n} \sum_{j=1}^n \operatorname{NLR}_j(\mathbf{X})
\end{align}
For our theoretical results, we restrict $\operatorname{MLP}_\theta$ to at least 1 hidden layer and width at least 2 with ReLU activations. 
\end{itemize}

Further details on how the definitions are applied and empirical data is collected are laid out in Appendix \ref{app:redundancy_measures}. In our empirical analyses, we effectively use an MLP with two hidden layers of dimensions 128 and 64 respectively.

%We constructively show that any linear function $\R^n \to \R$ can be captured by a two hidden layer MLP

\subsection{Mathematical Properties and Relationships}
\label{sec:proofs_section}

Given the above definitions, we now show that they are all bounded between 0 and 1, and derive how they are theoretically related to one another. The results rely on the assumption that all random variables are centered and have unit variance. 

\begin{lemmaE}[$0 \leq \operatorname{AAC}_j(\mathbf{X}) \leq \sqrt{\operatorname{LR}_j(\mathbf{X})} \leq \sqrt{\operatorname{NLR}_j(\mathbf{X})} \leq 1$][end, restate]
\label{lemma:mainInequality}
    For random variables $\mathbf{X} = [\textnormal{X}_1,\ldots \textnormal{X}_n]^T$ with $\Var[\textnormal{X}_i] = 1$ and $\E[\textnormal{X}_i] = 0$, for all $i \in [1,n]$ it holds that
    \begin{itemize}
        \item $\operatorname{AAC}_j(\mathbf{X})
        $, $\operatorname{LR}_j(\mathbf{X})$, and $\operatorname{NLR}_j(\mathbf{X})$ are bounded between 0 and 1.
        \item $\operatorname{AAC}_j(\mathbf{X}) \leq \sqrt{\operatorname{LR}_j(\mathbf{X})}$
         \item ${\operatorname{LR}_j(\mathbf{X})} \leq {\operatorname{NLR}_j(\mathbf{X})}$
    \end{itemize}
    % $0 \leq \operatorname{AAC}_j(\mathbf{X}) \leq \sqrt{\operatorname{LR}_j(\mathbf{X})} \leq \sqrt{\operatorname{NLR}_j(\mathbf{X})} \leq 1$.
\end{lemmaE}
\begin{proofE}
    We show the individual inequalities separately:

    \underline{$(0 \leq  \operatorname{AAC}_j(\mathbf{X}))$:} 
    By definition of $\operatorname{AAC}_j(\mathbf{X})$ we have 
\begin{align}
    \operatorname{AAC}_j(\mathbf{X}) &= \frac{\sum_{i= 1 :  i\neq j}^n |\Cov(\textnormal{X}_i, \textnormal{X}_j)|}{(n-1)} \geq 0
\end{align} as any sum over absolute values is greater or equals 0.

    \underline{$(\operatorname{AAC}_j(\mathbf{X}) \leq \sqrt{\operatorname{LR}_j(\mathbf{X})})$:} The equivalent $\operatorname{AAC}_j(\mathbf{X})^2 \leq \operatorname{LR}_j(\mathbf{X})$ follows from 
    \begin{align}
        \operatorname{LR}_j(\mathbf{X}) &= 1 - \min_{\vec{b}}  \E \left[|| \mathbf{X}_{-j} \cdot \vec{b} - \mathbf{X}_j ||_2 ^2 \right] = 1 - \min_{\vec{b}} \E \left[\left(\sum^n_{i=1, i\neq j} \left(\textnormal{X}_i b_i \right) - \textnormal{X}_j \right)^2\right] \\
        &= 1 - \min_{\vec{b}}\left( \sum_{i=1, i\neq j}^n \sum_{l=1, l\neq j}^n \Cov(\textnormal{X}_i,\textnormal{X}_l) b_i b_l - 2 \sum_{i=1, i\neq j}^n \Cov(\textnormal{X}_i,\textnormal{X}_j)b_i +  \Var[\textnormal{X}_j]\right)
    \end{align}
Let $k := \operatorname{argmax}_{1 \leq k \leq n, k \neq j} |\Cov(\textnormal{X}_k,\textnormal{X}_j)|$.
we now arbitrarily choose $b_l = \begin{cases}
  |\Cov(\textnormal{X}_k,\textnormal{X}_j)|  & \text{l = k} \\
  0 &  \text{else}
\end{cases}$ which allows
    \begin{align}
 &1 - \min_{\vec{b}}\left( \sum_{i=1, i\neq j}^n \sum_{l=1, l\neq j}^n \Cov(\textnormal{X}_i,\textnormal{X}_l) b_i b_l - 2 \sum_{i=1, i\neq j}^n \Cov(\textnormal{X}_i,\textnormal{X}_j)b_i +  \Var[\textnormal{X}_j]\right) \\
 &\geq -   \underbrace{\Cov(\textnormal{X}_k,\textnormal{X}_k)}_{\Var[\operatorname{X}_k] = 1} |\Cov(\textnormal{X}_k,\textnormal{X}_j)|^2 + 2 \Cov(\textnormal{X}_k,\textnormal{X}_j)|\Cov(\textnormal{X}_k,\textnormal{X}_j)| = \Cov(\textnormal{X}_k,\textnormal{X}_j) ^2
    \end{align}
    Meanwhile we also know that
    \begin{align}
\operatorname{AAC}_j(\mathbf{X})^2 &= \left(\frac{\sum_{i= 1 :  i\neq j}^n |\Cov(\textnormal{X}_i, \textnormal{X}_j)|}{(n-1)}\right)^2 \leq \left(\max_{1\leq i \leq n, i\neq j} \Cov(\textnormal{X}_i,\textnormal{X}_j)\right)^2
\end{align}
combining the two inequalities derived above yields $(\operatorname{AAC}_j(\mathbf{X}) \leq \sqrt{\operatorname{LR}_j(\mathbf{X})})$ as desired.

\underline{$({\operatorname{LR}_j(\mathbf{X})} \leq {\operatorname{NLR}_j(\mathbf{X})})$:}  

We show that any linear map $f : \R^n \to \R$ with $f(\vec{x}) =  \vec{x} \cdot \vec{b}$, where $\vec{b}$ is the vector of coefficients, can also be represented by an MLP using ReLU with at least one hidden layer with width at least 2.

For the following discussion, we set all biases to 0. Let $h_{i,j}$ be the value at neuron $j$ in hidden layer $i$ after the application of the activation function. 

By setting $W_{h_{1,1}} = \beta$ and $W_{h_{1,2}} = -\beta$, where $W_{h}$ is the weights' matrix for a neuron $h$ which is used for multiplying with incoming activations yields 
\begin{itemize}[]
    \item $h_{1,1} = \max \left(\vec{x}  \cdot\vec{b},0 \right) $
    \item $h_{1,2} = \max \left( -\vec{x}  \cdot \vec{b},0 \right) $.
\end{itemize}

For any following layer except the final regressive layer we set $W_{h_{i\geq 2,1}} = [1,0]$ and $W_{h_{i\geq 2,2}} = [0,1]$ yielding

\begin{itemize}[]
    \item $h_{i,1} = \max \left(\vec{x}  \cdot\vec{b},0 \right) $
    \item $h_{i,2} = \max \left( -\vec{x}  \cdot \vec{b},0 \right) $.
\end{itemize}

For the final regressive layer we use weights [1,-1] which yields a final output of \begin{align}
    \max \left(\vec{x}  \cdot\vec{b},0 \right) -  \max \left( -\vec{x}  \cdot \vec{b},0 \right) = \vec{x}  \cdot \vec{b}
\end{align} as desired. This shows that any MLP with at least one hidden layer and width 2 can perfectly reconstruct any linear map 

For an MLP with width larger than 2 we choose an arbitrary subset of two neurons per layer to embed our constructed MLP from above and set all other weights to 0.

Hence, any linear mapping can be represented by an MLP with width at least 2 and at least 1 hidden layer. From this and the definition from NLP it follows that $\min_b || \mathbf{X}_{-j} \cdot b - \mathbf{X}_j ||_2 ^2 \geq \min_\theta || R_\theta\left(\mathbf{X}_{-j} \right) - \mathbf{X}_j ||_2 ^2 $ as desired. 

\underline{$({\operatorname{NLR}_j(\mathbf{X})} \leq 1)$:} Clearly $\min_\theta || R_\theta\left(\mathbf{X}_{-j} \right) - \mathbf{X}_j ||_2 ^2 \geq 0$ as the square delimits the expression from below at 0. From this it directly follows that $\operatorname{NLR}_j(\mathbf{X}) = 1 - \min_\theta \E\left[ || R_\theta\left(\mathbf{X}_{-j} \right) - \mathbf{X}_j ||_2 ^2 \right] \leq 1$.

From the proven individual inequalities, the claimed results follow. 
\end{proofE}

\begin{theoremE}[$0 \leq \operatorname{AAC}(\mathbf{X}) \leq \sqrt{\operatorname{LR}(\mathbf{X})} \leq \sqrt{\operatorname{NLR}(\mathbf{X})} \leq 1$][end, restate]\label{thm:mainTheorem}
  For random variables $\mathbf{X} = [X_1,\ldots X_n]^T$ with $\Var[X_i] = 1$ and $\E[X_i] = 0$ for all $i \in [1,n]$ it holds that 
    \begin{itemize}
        \item $\operatorname{AAC}(\mathbf{X})
        $, $\operatorname{LR}(\mathbf{X})$, and $\operatorname{NLR}(\mathbf{X})$ are bounded between 0 and 1.
        \item $\operatorname{AAC}(\mathbf{X}) \leq \sqrt{\operatorname{LR}(\mathbf{X})}$
         \item ${\operatorname{LR}(\mathbf{X})} \leq {\operatorname{NLR}(\mathbf{X})}$
    \end{itemize}
\end{theoremE}
\begin{proofE}

This follows directly applying Lemma \ref{lemma:mainInequality} $n$ times (for every summand) and using the fact that we can upper bound an arithmetic mean by a quadratic mean. 

\begin{align}
    \operatorname{AAC}(\mathbf{X}) = \frac{1}{n} \sum_{j=1}^n \operatorname{AAC}_j(\mathbf{X}) \geq  \frac{1}{n} \sum_{j=1}^n 0 = 0
\end{align}
\begin{align}
    \operatorname{AAC}(\mathbf{X}) = \frac{1}{n} \sum_{j=1}^n \operatorname{AAC}_j(\mathbf{X}) \leq \frac{1}{n} \sum_{j=1}^n \sqrt{\operatorname{LR}_j(\mathbf{X})} \leq \sqrt{\frac{1}{n}  \sum_{j=1}^n \operatorname{LR}_j(\mathbf{X}) } = \operatorname{LR}(\mathbf{X})
\end{align}
\begin{align}
    \operatorname{LR}(\mathbf{X}) =  \frac{1}{n} \sum_{j=1}^n \operatorname{LR}_j(\mathbf{X})  \leq \frac{1}{n} \sum_{j=1}^n \operatorname{NLR}_j(\mathbf{X}) = \operatorname{NLR}(\mathbf{X})
\end{align}

Combining the inequality yields the desired result. 

\end{proofE}

\begin{corollaryE}[][end, restate]
\label{corr:iszero}
For $\textnormal{X}_1, \ldots \textnormal{X}_n$ mutually independent variables with all $\textnormal{X}_i$ centered unit variance random variables, it follows that $0 = \operatorname{AAC}(\mathbf{X}) = \sqrt{\operatorname{LR}(\mathbf{X})} = \sqrt{\operatorname{NLR}(\mathbf{X})}$
\end{corollaryE}
\begin{proofE}
By Lemma \ref{lemma:mainInequality} and Theorem \ref{thm:mainTheorem} it suffices to show that if $\textnormal{X}_1, \ldots \textnormal{X}_n$ are mutually independent variables with all $\textnormal{X}_i$ centered unit variance that $\operatorname{NLR}_j(\mathbf{X}) = 0$ for all $j$.
This follows from 
\begin{align}
   \operatorname{NLR}_j(\mathbf{X})  &=   1 - \min_\theta \E \left[  || R_\theta\left(\mathbf{X}_{-j} \right) - \textnormal{X}_j ||_2 ^2 \right]\\
   &= 1 - \min_\theta \left( \E\left[ R_\theta (\mathbf{X}_{-j})^2\right] -2 \E \left[ R_\theta (\mathbf{X}_{-j})\textnormal{X}_j \right] + \Var[\textnormal{X}_j] \right)\\
   &= - \min_\theta \left( \E\left[ R_\theta (\mathbf{X}_{-j})^2\right] \right) = 0
\end{align}
where we have used the assumed independence at $\E \left[ R_\theta (\mathbf{X}_{-j})\textnormal{X}_j \right] = \E \left[ R_\theta (\mathbf{X}_{-j})\right] \E \left[\textnormal{X}_j \right]$ and minimized the expression by using the predictor $R_0(\mathbf{X}) = 0$. As this holds for any $j$ we conclude that $\operatorname{NLR}(\mathbf{X}) = 0$ from which the corollary follows.
\end{proofE}

\begin{corollaryE}[][end, restate]
\label{corollary:zero}
    For $\textnormal{X}_1,\ldots,\textnormal{X}_n$ centered random variables with unit variance, it holds that
\begin{align}
    \operatorname{AAC}([\textnormal{X}_1,\ldots,\textnormal{X}_n]) = 0 \iff \operatorname{LR}([\textnormal{X}_1,\ldots,\textnormal{X}_n]) = 0
\end{align}
    
\end{corollaryE}

\begin{proofE}
    \underline{($\impliedby$)}

    By Theorem \ref{thm:mainTheorem} if $\operatorname{LR}([\textnormal{X}_1,\ldots,\textnormal{X}_n]) = 0$ then so must $\operatorname{AAC}([\textnormal{X}_1,\ldots,\textnormal{X}_n]) = 0$.

    \underline{($\implies$)}
Given $\operatorname{AAC}([\textnormal{X}_1,\ldots,\textnormal{X}_n]) = 0$ all pairwise covariances $\Cov ( \textnormal{X}_i,\textnormal{X}_j)$ must be zero. 

\begin{align}
   \operatorname{LR}_j([\textnormal{X}_1,\ldots,\textnormal{X}_n])  &=   1 - \min_{\vec{b}} \E \left[  || \sum_{i=1, i \neq j }^n b_i \textnormal{X}_i - \textnormal{X}_j ||_2 ^2 \right]\\
   &= 1 - \min_{\vec{b}} \left( \E\left[ \sum_{i=1, i \neq j }^n \sum_{k=1, k \neq j }^n b_i b_k \textnormal{X}_i\textnormal{X}_k\right] -2 \E \left[\sum_{i=1, i \neq j }^n b_i \textnormal{X}_i  \textnormal{X}_j \right] + \Var[\textnormal{X}_j] \right)\\
   &=  - \min_{\vec{b}} \left( \sum_{i=1, i \neq j }^n \sum_{k=1, k \neq j }^n b_i b_k \E\left[ \textnormal{X}_i\textnormal{X}_k\right] -2 \sum_{i=1, i \neq j }^n b_i \E \left[\textnormal{X}_i  \textnormal{X}_j \right] \right)\\
   &= - \min_{\vec{b}} \left(  \sum_{i=1, i \neq j }^n b_i^2 \E\left[ \textnormal{X}_i^2\right]  \right)=  - \min_{\vec{b}} \left(  \sum_{i=1, i \neq j }^n b_i^2 \cdot 1 \right) = 0
\end{align}

as $\E[\textnormal{X}_i\textnormal{X}_j] = 0$ due to $\Cov ( \textnormal{X}_i,\textnormal{X}_j) = 0$ and the variables being centered. To minimize, we choose $\vec{b} = \vec{0}$.
As $j$ was arbitrary, this holds for any $\operatorname{LR}_j([\textnormal{X}_1,\ldots,\textnormal{X}_n])$ and hence $\operatorname{LR}([\textnormal{X}_1,\ldots,\textnormal{X}_n]) = 0$ as desired.
\end{proofE}

Informally speaking by Corollary \ref{corollary:zero}, $\operatorname{AAC}$ and $\operatorname{LR}$ are similar in terms of the redundancies they can maximally capture, which begs the question of why we distinguish between them.

For this, let us look at the example where $\textnormal{X}_1, \ldots, \textnormal{X}_n$ are centered random variables with unit variance with $\textnormal{X}_n = \frac{1}{\sqrt{n-1}} \textnormal{X}_1 +  \ldots + \frac{1}{\sqrt{n-1}} \textnormal{X}_{n-1}$ but with $\textnormal{X}_1, \ldots \textnormal{X}_{n-1}$ mutually independent. 

As any variable is completely linearly reconstructable from all other ones, it follows that 
$\operatorname{LR}([\textnormal{X}_1, \ldots, \textnormal{X}_n]) = \operatorname{NLR}([\textnormal{X}_1, \ldots, \textnormal{X}_n]) = 1$.

However, using independence of the first $n-1$ variables and
\begin{align}
     \Cov(\textnormal{X}_i, \textnormal{X}_n) &= \Cov \left(\textnormal{X}_i, \frac{1}{\sqrt{n-1}} \textnormal{X}_1 +  \ldots + \frac{1}{\sqrt{n-1}} \textnormal{X}_{n-1}\right) = \frac{\Var[\textnormal{X}_i]}{\sqrt{n-1}}  = \frac{1}{\sqrt{n-1}}
\end{align}

we find 
\begin{align}
    \operatorname{AAC}([\textnormal{X}_1, \ldots, \textnormal{X}_n]) = \frac{0 \cdot \binom{n-1}{2} + \sum_{i=1}^{n-1} \Cov(\textnormal{X}_i, \textnormal{X}_n)}{\binom{n}{2}} = \frac{\sqrt{n-1}}{\binom{n}{2}} \in \mathcal{O}(n^{-3/2})
\end{align}    
meaning that $ \operatorname{AAC}([\textnormal{X}_1, \ldots, \textnormal{X}_n])$ becomes diminishingly small as $n$ increasing.

Hence, whereas $\operatorname{AAC}$ measures how strongly variables are \emph{pairwise} correlated, $\operatorname{LR}$ and $\operatorname{NLR}$ intuitively speaking 
also capture interactions involving three or more variables. They measure to what degree information about each variable is contained within the others.

Furthermore, the nonlinear nature of $\operatorname{NLR}$ allows it to capture more redundancy than $\operatorname{LR}$ as Corollary \ref{corr:NLRMorePower} demonstrates.

\begin{corollaryE}[][end, restate]
\label{corr:NLRMorePower}
 For $\textnormal{X}_1,\ldots,\textnormal{X}_n$ centered variables with unit variance, $\operatorname{LR}([\textnormal{X}_1,\ldots,\textnormal{X}_n]) = 0 \centernot\implies  \operatorname{NLR}([\textnormal{X}_1,\ldots,\textnormal{X}_n]) = 0$
\end{corollaryE}
\begin{proofE}
    We show a counterexample which proves the implication cannot hold. Let $\textnormal{X},\textnormal{Y} \sim \mathcal{N}(0,1)$. Further, let $\textnormal{Z} = \frac{1}{2}\left(\textnormal{X}^2 - \textnormal{Y}^2\right)$. We have 
    \begin{align}
        \E[\textnormal{Z}] &= \E\left[\frac{1}{2}\left(\textnormal{X}^2 - \textnormal{Y}^2\right)\right] = \frac{1}{2} \left( \E[\textnormal{X}^2] - \E[\textnormal{Y}^2]\right) = 0\\
        \Var[\textnormal{Z}] &= \Var\left[\frac{1}{2}\left(\textnormal{X}^2 - \textnormal{Y}^2\right)\right] = \frac{1}{4} \left(\Var\left[\textnormal{X}^2\right]  + \Var\left[ \textnormal{Y}^2\right]\right) = \frac{1}{4} \left( 3 - 1 + 3 - 1\right) = 1
    \end{align}
where we used the fact that the 4-th moment of a centered unit variance Gaussian is $3$ and the second moment is $1$.

For $\operatorname{NLR}([\textnormal{X}, \textnormal{Y},\textnormal{Z}])$ we find 
{\footnotesize
\begin{align}
    \operatorname{NLR}([\textnormal{X}, \textnormal{Y},\textnormal{Z}]) &= \frac{3 -  \min_\theta \E \left[  || R_\theta\left(\textnormal{X}, \textnormal{Y}\right) - \textnormal{Z} ||_2 ^2 \right] - \min_\theta \E \left[  || R_\theta\left(\textnormal{X}, \textnormal{Z}\right) - \textnormal{Y} ||_2 ^2 \right] - \min_\theta \E \left[  || R_\theta\left(\textnormal{Z}, \textnormal{Y}\right) - \textnormal{X} ||_2 ^2 \right] }{3} = 1
\end{align}
}
as given any two variables, a nonlinear predictor is able to perfectly reconstruct the third variable by construction of our example.

Given by assumption $\operatorname{LR}([\textnormal{X}_1,\ldots,\textnormal{X}_n]) = 0$ we must by Corollary \ref{corollary:zero} also have $\operatorname{AAC}([\textnormal{X}_1,\ldots,\textnormal{X}_n]) = 0$ from which we infer
{\footnotesize
\begin{align}
    \operatorname{AAC}([\textnormal{X},\textnormal{Y},\textnormal{Z}]) &= \frac{|\Cov\left(  \textnormal{X},\textnormal{Y} \right)| + |\Cov\left(  \textnormal{X},\textnormal{Z} \right)| + |\Cov\left(  \textnormal{Y},\textnormal{Z} \right)|}{3} = \frac{|\Cov\left(  \textnormal{X},\frac{1}{2}\left(\textnormal{X}^2 - \textnormal{Y}^2\right) \right)| + |\Cov\left(  \textnormal{Y},\frac{1}{2}\left(\textnormal{X}^2 - \textnormal{Y}^2\right)\right)|}{3}\\
&= \frac{|\Cov\left(  \textnormal{X},\textnormal{X}^2 - \textnormal{Y}^2\right)| + |\Cov\left(  \textnormal{Y},\textnormal{X}^2 - \textnormal{Y}^2\right)|}{12}\\
=& \frac{|\Cov\left(  \textnormal{X},\textnormal{X}^2 \right)| + |\Cov\left(  \textnormal{X}, \textnormal{Y}^2\right)| + |\Cov\left(  \textnormal{Y},\textnormal{X}^2\right)| + |\Cov\left(  \textnormal{Y},\textnormal{Y}^2\right)|}{12} = 0
\end{align}
}
where we have used $\Cov(\textnormal{X},\textnormal{Y}) = 0$ as well as $\Cov(\textnormal{X},\textnormal{Y}^2) = 0$ which follow from the independence of the centered Gaussians $X$ and $Y$. Further $\Cov(\textnormal{X}^2,\textnormal{X}) = \E[\textnormal{X}^3] - \E[\textnormal{X}]\E[\textnormal{X}^2] = 0$.

Which shows that although $\operatorname{LR}$ and $\operatorname{NLR}$ are zero $\operatorname{NLR}$ takes its maximal value for this example. This proves the Corollary by counterexample.
\end{proofE}

\subsection{SSLPM}

To understand the role of redundancies in SSL, we introduce Self Supervised Learning with Predictability Minimization (SSLPM) where instead of reducing pairwise correlations as in Barlow Twins (Equation \ref{eq:barlowTwinsLoss}) we minimize redundancy via predictability minimization \citep{schmidhuber1996semilinear}. We do this by using a standard Siamese set-up for SSL but extended with a predictor network operating on the concatenated representations from the two augmentations. The loss of the encoder is then given by the invariance loss combined with the new prediction loss coming from the predictor. 

\begin{figure}[h]
\begin{center}
\includegraphics[width=1\textwidth]{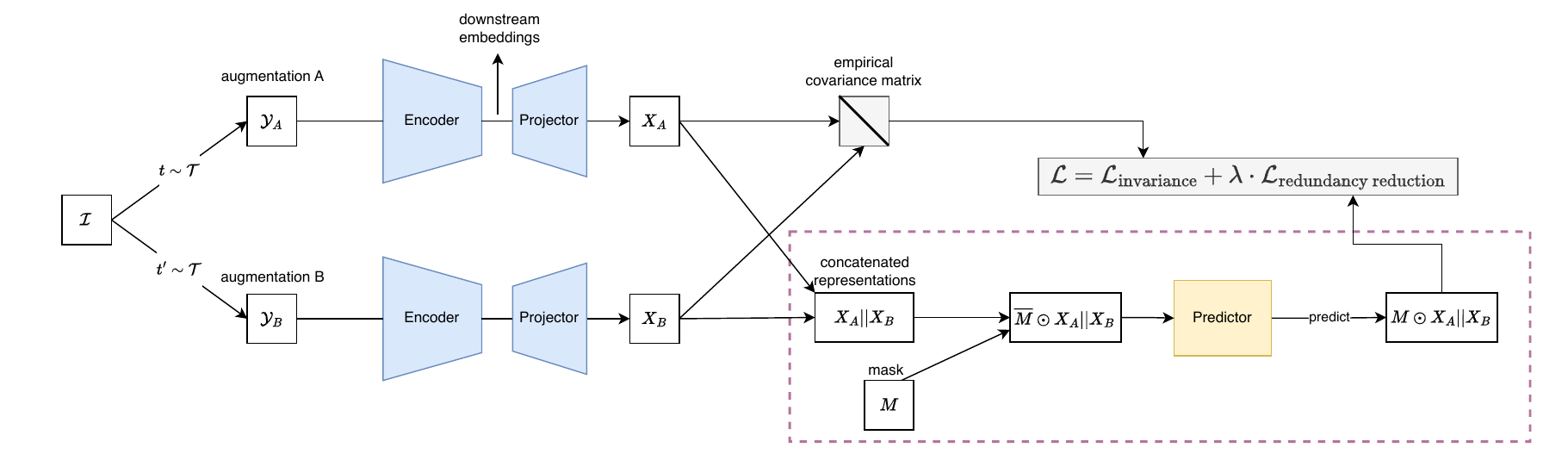}
\end{center}
\caption{Schematic representation of the SSLPM model with the two actors, the encoder-projector network (blue) and the predictor network (yellow). Our contribution is indicated with the dashed box.}
\label{fig:main_model_diagram}
\end{figure}

The set-up of SSLPM is illustrated in Figure \ref{fig:main_model_diagram}.

\paragraph{Predictability Minimization}

Since representations can be high-dimensional, we depart from the typical leave-one-out predictability minimization scheme introduced by \citet{schmidhuber1996semilinear} 
% and used by \citet{schraudolph1999processing} 
and take inspiration from random masking \cite{devlin2018bert, hsu2021hubert} originally introduced in language modeling. Instead of performing $n$ leave-one-out predictions, we train a single network to predict randomly masked input features. 

For this, let us define $\mathcal{B}$ as a mini batch of $b = |\mathcal{B}|$ images that are fed through the encoder and projector, resulting (after standardization) in two augmented representations, $\mathbf{X_A}$ and $\mathbf{X_B} \in \R^{b\times n}$.

After concatenating them into a single tensor $\mathbf{X_{A||B}} \in \R^{2b \times n}$, we generate a binary mask $\mathbf{M} \in \R^{2b\times n}$ with $\mathbf{M}_{i,j}\sim \operatorname{Bernoulli}(1/2)$, i.e., in expectation half of all input features are masked. In the following, we further denote the inverse mask by $\mathbf{\Bar{M}} := 1 - \mathbf{M}$.

Let $R_\theta : \R^n \to \R^n$ be a predictor parametrized by $\theta$. We aim to train $R_\theta$ such that for a randomly sampled mask $\mathbf{M}$, we minimize the average squared prediction error of the masked entries. This can be expressed as the following minimization problem:
\begin{align}
\label{eq:predictorMinimzation}
&\min {\sum_{i,j} \mathbf{M}_{i,j}  \left(R_\theta(\mathbf{\Bar{M}} \odot {\mathbf{X_{A||B}}})_{i,j} - {\mathbf{X_{A||B}}}_{i,j}\right)^2}\\ &= \min \norm{\mathbf{M} \odot \left(R_\theta(\mathbf{\Bar{M}} \odot {\mathbf{X_{A||B}}}) - {\mathbf{X_{A||B}}}\right)}_F^2,
\end{align}

where $\odot$ denotes elementwise multiplication. 
In other words, we aim to predict the masked indices of the concatenated representations from the non-masked indices. Note that we only care about the indices that were masked, reconstruction of the others does not contribute to the loss.

Combining this with the invariance loss used in Barlow Twins (Equation \ref{eq:barlowTwinsLoss}) yields the following loss function for SSLPM:
\begin{align}\label{eq:clLoss}
    \mathcal{L}_{\text{SSLPM}} &= \underbrace{\sum _{i} \left(1- \left(\mathbf{X_A} ^T \cdot \mathbf{X_B}\right)_{ii}\right)^2}_{\mathcal{L}_\text{invariance}}   -\lambda \cdot \underbrace{\frac{1}{|\mathbf{M}|}{\sum_{i,j} \mathbf{M}_{i,j}  \left(R_\theta({\mathbf{X_{A||B}}})_{i,j} - {\mathbf{X_{A||B}}}_{i,j}\right)^{2},} }_{\mathcal{L}_{\text{pred}}}
\end{align}
where the prediction loss is adjusted for the number of masked entries $|\mathbf{M}| = \sum_{i,j}\mathbf{M}_{i,j}$, and the relative importance of the two loss terms is controlled by the hyperparameter $\lambda$.

\paragraph{SSLPM: A Competitive Game}
By construction, the two networks, the encoder and the predictor, engage in a competitive game. The predictor aims to predict the masked features and the encoder aims to 
(1) ensure corresponding features from different augmentations correlate with each other and 
(2) make the predictor's task as hard as possible. This is similar to a GAN \citep{goodfellow2014generative} set-up, where the loss of the generator depends on the loss of the discriminator.

\paragraph{Predictor Design}

In the discussion above we have intentionally not put any constraints on the predictor architecture $R_\theta$. We now introduce the two predictor designs we use in our experiments:
\begin{itemize}
    \item SSLPM-SGD: Self supervised learning with stochastic gradient descent-based predictability minimization, where the predictor network is optimized using gradient descent, and
    \item SSLPM-RR: Self supervised learning with ridge regression-based predictability minimization, where a closed form solution for the optimal ridge regression weights is used. 
\end{itemize}

\subsubsection{SSLPM-SGD}

In the gradient descent-based predictor, we use an MLP predictor with two hidden layers (dimensions: 512-128-64-1) and ReLU activations. After every forward pass of the encoder, the standardized and merged representations $\mathbf{X_{A||B}}$ are used to train the predictor for multiple iterations. Note that the objective of the predictor (Equation \ref{eq:predictorMinimzation}) is dependent on the mask $\mathbf{M}$. To increase stability, we sample a new random mask for each optimization step. Additionally, we sample a validation mask $\mathbf{M}_V$ at the beginning, which is used for early stopping. Once the optimization of the predictor has concluded, $\mathbf{M}_V$ and the optimized predictor are used to compute the predictability loss for the encoder.

\subsubsection{SSLPM-RR}
If the predictor is a linear layer, we can solve for its optimal weights directly without gradient-based optimization.
We define the predictor network as $R_{\bm{W}}(X) = X \bm{W}$ where $\bm{W} \in \R^{k\times k}$ is the weight matrix of a multilinear regression.
%We deviate from the loss function presented in Equation \ref{eq:predictorMinimzation} and introduce a ridge penalty for the network weights using the following loss function for the predictor:
We deviate slightly from the loss function presented in Equation \ref{eq:predictorMinimzation} to one more suited to linear regression and introduce a ridge penalty for the network weights using the following loss function for the predictor:
\begin{align}
&\min \norm{R_{\bm{W}}(\mathbf{\Bar{M}} \odot {\mathbf{X_{A||B}}}) - {\mathbf{X_{A||B}}}}_F^2 + \mu \norm{{\bm{W}}}_F^2.
\end{align}
This leads to the following closed-form solution for the weights:
\begin{align}
\bm{W} = \left(\left(\left(\mathbf{\Bar{M}} \odot \mathbf{X_{A||B}}\right)^T \cdot \left(\mathbf{\Bar{M}} \odot \mathbf{X_{A||B}}\right) + \mu \mI \right)^{-1} \cdot \left(\mathbf{\Bar{M}} \odot \mathbf{X_{A||B}}\right)^T \cdot \mathbf{X_{A||B}} \right),
\end{align}
where $\mu$ is the ridge penalty. The ridge penalty is needed when the dimensionality of the regression is large but the batch size is small. In this case, linear regression would involve inverting a singular matrix, but introducing a Ridge penalty circumvents this problem. More details on the loss function as well as other design choices can be found in Appendices \ref{app:SSLPMREG} and \ref{app:further_analysis}.

\subsubsection{Implementation Details}
\label{sec:implementation_details}
Our implementation is based on \citet{tsai2021note} and implemented in the solo-learn \citep{da2022solo} framework. For experimental consistency, all models were (re-)trained on our infrastructure. 

For all models, except SSLPM, we use the best performing hyperparameters from the solo-learn repository, which the authors optimized extensively for CIFAR-10 and ImageNet-100. 

SSLPM-SGD was optimized on CIFAR-10 with the hyperparameters then also being used on CIFAR-100 and ImageNet-100. SSLPM-RR was optimized on CIFAR-10 and ImageNet-100 in line with the models from solo-learn. Further implementation details can be found in Appendix \ref{app:implementation_details}.

\section{Results \& Analysis}
\label{sec:main_results}

In the following, we perform a deeper analysis into embedding space redundancies and an investigation into SSLPM. As SSLPM-RR outperforms SSLPM-SGD in our experiments, most analyses are based on the former. 

\subsection{Overall Performance}

For all methods used in our work, we report accuracy numbers on the three datasets: CIFAR-10, CIFAR-100, and ImageNet-100. For the CIFAR-10 and CIFAR-100 datasets, we use a ResNet-18 backbone \citep{he2016deep} trained for 1000 epochs and report Top 1 and Top 5 accuracies.
We use the same methodology as solo-learn \citep{da2022solo} with an on-line linear prediction head that is jointly trained on top of the ResNet-18 backbone during the backbone training. 
For ImageNet-100 we use a ResNet-18 backbone trained for 400 epochs and report the same Top 1 and Top 5 accuracies. In appendix \ref{app:offline_vs_online} additional results for offline evaluation on ImageNet100 are reported.

Barlow Twins, SSLPM-RR, and SSLPM-SGD were trained with three different seeds, and mean scores with standard deviations are reported. For SSLPM-SGD we present SSLPM-SGD 1l results using a one-layer MLP predictor and SSLPM-SGD 3l results using a three-layer MLP predictor. Details on hyperparameters and training can be found in Appendix \ref{app:implementation_details}.

\begin{table}[h]
\setlength{\tabcolsep}{5pt}
\centering
\caption{Evaluation of different models on various datasets. \textbf{\underline{Top1}}, \textbf{Top2}, and \underline{Top3} models are highlighed per dataset.}
\label{table:resModels}
\footnotesize
\begin{tabular}{@{}lcc|cc|cc@{}}
\toprule
             & \multicolumn{2}{c}{CIFAR-10} & \multicolumn{2}{c}{CIFAR-100} & \multicolumn{2}{c}{ImageNet-100} \\
\cmidrule(lr){2-3} \cmidrule(lr){4-5} \cmidrule(lr){6-7} 
Models       & Top 1         & Top 5        & Top 1         & Top 5         & Top 1           & Top 5      \\
\midrule
Barlow Twins &  \underline{92.21 (0.136)} & \textbf{99.82 (0.026)} & \textbf{70.94 (0.075)} & \textbf{92.35 (0.042)}  & \textbf{\underline{80.31 (0.099)}} & \textbf{\underline{95.33 (0.147)}}      \\
BYOL         & \textbf{\underline{93.03}} & 99.76 &  \textbf{\underline{71.04}} &  \textbf{\underline{92.42}}  & 76.12 &   93.92     \\
NNCLR        & 91.70 & 99.76 & 68.61 & 91.34 & \underline{79.22} & 94.82 \\
SimCLR       & 91.23 & 99.76 & 66.63 & 89.58 & 77.38 & 94.02  \\
MocoV3       & 90.40 & 99.77 & 66.94 & 90.21 & 74.72 & 93.02  \\
VICReg       & 91.71 & 99.71 & 68.23 & 90.86 & \textbf{79.56} & \textbf{95.10}  \\
VIbCReg      & 90.51 & \underline{99.80} & 66.22 & 89.50 & 78.14 & 94.02 \\
\textbf{SSLPM-RR (ours)} & \textbf{92.70 (0.144)}   & \textbf{\underline{99.85 (0.029)}} &  \underline{70.46 (0.230)} & \underline{92.14 (0.196)} &    79.17 (0.310)    &   \underline{95.02 (0.178)}   \\
\textbf{SSLPM-SGD 1l} & 90.39 (0.136)	& 99.72 (0.015) & 67.74 (0.083)	& 90.70 (0.240)& 76.19 (0.196)	& 93.90 (0.193)	   \\
\textbf{SSLPM-SGD 3l} & 86.69 (0.295) & 99.62 (0.031) & 47.65 (0.201) & 76.59 (0.170) & 64.27 (0.594)	&88.55 (0.600)	 \\
\bottomrule
\end{tabular}
\end{table}

Table \ref{table:resModels} shows most previous methods performing similarly well, with MocoV3 falling slightly behind. We also see that SSLPM-RR is competitive with Barlow Twins and other SOTA SSL methods. However, SSLPM-SGD with one MLP layer, although being highly similar to SSLPM-RR, falls a few percentage points behind. When the number of layers in SSLPM-SGD is increased from 1 to 3, the gap widens.

\begin{table}[h]
\centering
\caption{Ablation showing the importance of the redundancy reduction term.}
\label{table:lambda_ablation}
\footnotesize
\begin{tabular}{@{}lcc|cc|cc@{}}
\toprule
             & \multicolumn{2}{c}{CIFAR-10} & \multicolumn{2}{c}{CIFAR-100} & \multicolumn{2}{c}{ImageNet-100} \\
\cmidrule(lr){2-3} \cmidrule(lr){4-5} \cmidrule(lr){6-7} 
Models       & Top 1         & Top 5        & Top 1         & Top 5         & Top 1           & Top 5        \\
\midrule
{SSLPM-RR} & {92.70 (0.144)}   & 99.85 (0.029) &  70.46 (0.230) & 92.14 (0.196) &    79.17 (0.310)    &   95.02 (0.178)   \\
{SSLPM-SGD 1l (ours)} & 90.39 (0.136)	& 99.72 (0.015) & 67.74 (0.083)	& 90.70 (0.240) & 76.19 (0.196)	& 93.90 (0.193)	   \\
{SSLPM-SGD 3l (ours)} & 86.69 (0.295) & 99.62 (0.031) & 47.65 (0.201) & 76.59 (0.170) & 64.27 (0.594)	&88.55 (0.600)\\
SSLPM ($\lambda = 0$) &   77.71  (0.106) &  98.95 (0.060)    &  34.05 (1.081)    &  63.74 (1.358)      &    50.81 (0.388)      &   79.86 (0.046)   \\
\bottomrule
\end{tabular}
\end{table}
To quantify the impact of redundancy reduction, we trained a Barlow Twins model without redundancy reduction ($\lambda = 0$). This essentially yields a SSLPM base model which has accuracies significantly lower than any proposed models with redundancy reduction, as can be seen in Table \ref{table:lambda_ablation}.

These results show that:
\begin{enumerate}
    \item More general forms of redundancy reduction can work just as well as the covariance-based redundancy reduction used in Barlow Twins.
    \item Reducing higher-order redundancies (e.g. SSLPM-SGD-3l) can result in lower performance, though still outperforming the $\lambda = 0$ baseline.
\end{enumerate}

\subsection{Redundancy versus Performance in SSLPM}

\begin{figure}[h]
  \centering
  \includegraphics[width=0.7\textwidth]{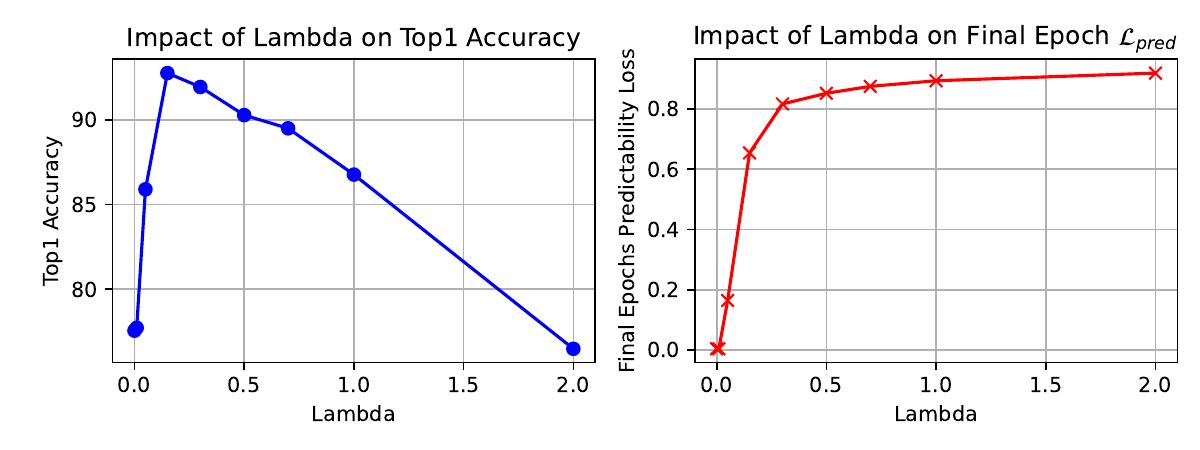}
  \caption{Impact of $\lambda$ in Equation \ref{eq:clLoss} on CIFAR-10.}
\label{fig:lambda_vs_top1_accuracy_and_predictability_loss}
\end{figure}

The hyperparameter $\lambda$ as introduced in Equation \ref{eq:clLoss} plays a crucial role in weighing the different losses against one another. Figure \ref{fig:lambda_vs_top1_accuracy_and_predictability_loss} shows final Top-1 accuracy in the left plot and the predictability loss in the final epoch of training in the right plot. Although we see increasing $\mathcal{L}_{pred}$ for increasing $\lambda$, optimal performance is reached before maximal predictability loss is reached and falls sharply for $\lambda > 0.25$. More detailed plots on how this evolves during training can be found in Appendix \ref{app:impact_of_lamba_on_training}.

The observations from Figure \ref{fig:lambda_vs_top1_accuracy_and_predictability_loss} indicate that too much redundancy reduction reduces embedding quality. However, stronger emphasis on redundancy reduction in the loss must not result in less redundancy in the final embeddings. 

\begin{figure}[h]
  \centering
  \includegraphics[width=\textwidth]{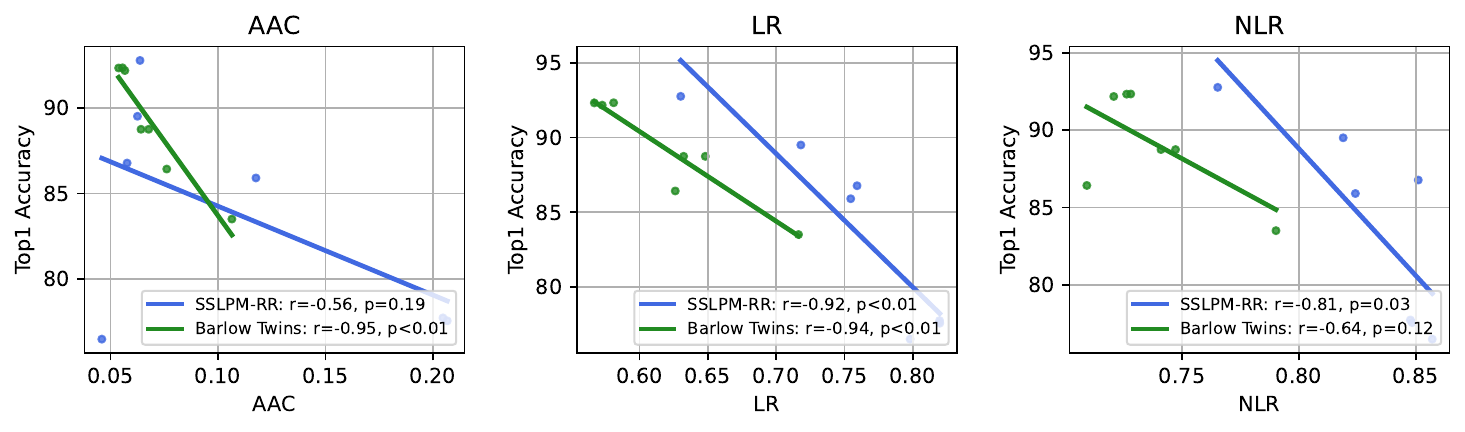}
  \caption{Relationship between Top-1 accuracy and three different redundancy measures on CIFAR-10 plotted with the lines of best fit. For the best fit lines, the Pearson correlation and the $p$-value for the null hypothesis that the data is uncorrelated is reported. LR is the only redundancy measure where the correlation is significant (with $p<0.01$) for both models.}
\label{fig:accuracy_redundancy_relationship_plot_cifar10}
\end{figure}

For this, given that different $\lambda$'s result in different performance, we set out to look at these results from a different angle: How does final model performance relate to the redundancy in the embedding space? In Figure \ref{fig:accuracy_redundancy_relationship_plot_cifar10}
we see that for LR there is a clear link, whereby less redundancy results in higher performance (with $p<0.01$ across all data sets and models). However, for AAC and NLR there seems to be an optimal level of redundancy after which performance degrades. Similar results can be seen in Figure \ref{fig:accuracy_redundancy_relationship_plot_cifar100_imagenet_100} in the appendix for the CIFAR-100 and ImageNet-100 datasets. 

From these results one can infer that reducing the redundancy in Barlow Twins and SSLPM-RR aids performance. However, this is not necessarily achieved by increasing the hyperparameter $\lambda$. At a certain point the redundancy reduction term is too strong compared to the invariance loss leading to a drop in performance.

\subsection{More Powerful Predictability Minimization}

\begin{figure}[h]
  \centering
  \includegraphics[width=\textwidth]{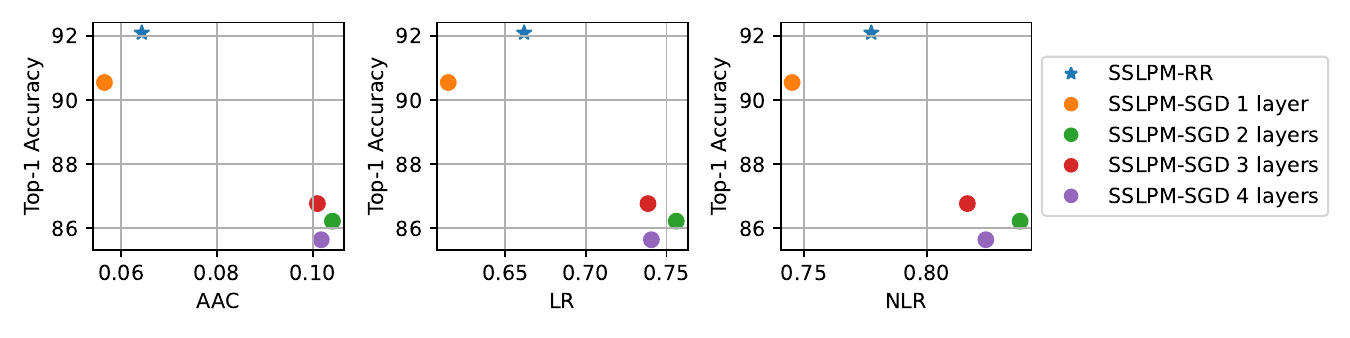}
  \caption{Ablation on the number of layers of the predictor in SSLPM-SGD compared with SSLPM-RR. All results are calculated on CIFAR-10.}
  \label{fig:layers_predictor_SSLPM_opt}
\end{figure}

In comparison to SSLPM-RR, where we have a fixed regression problem to solve in the predictor, we can freely design the predictor in SSLPM-SGD. In Figure \ref{fig:layers_predictor_SSLPM_opt} we see that (1) all SSLPM-SGD versions underperform SSLPM-RR in terms of accuracy, but the 1 layer SSLPM-SGD version features the least redundancy for all measures, 
and (2) when the predictor in SSLPM-SGD contains two or more layers, meaning that the predictor can capture non-linearities, there is a steep drop in accuracy as well as the redundancy that is removed. We therefore conclude that there is no evidence that reducing higher-order redundancies is beneficial for downstream embedding performance - in fact we see evidence to the contrary, with performance and actual redundancy being removed falling as the number of layers of the predictor increases. 

\subsection{Projector Depth and Redundancy}

\begin{figure}[h]
  \centering
  \includegraphics[width=\textwidth]{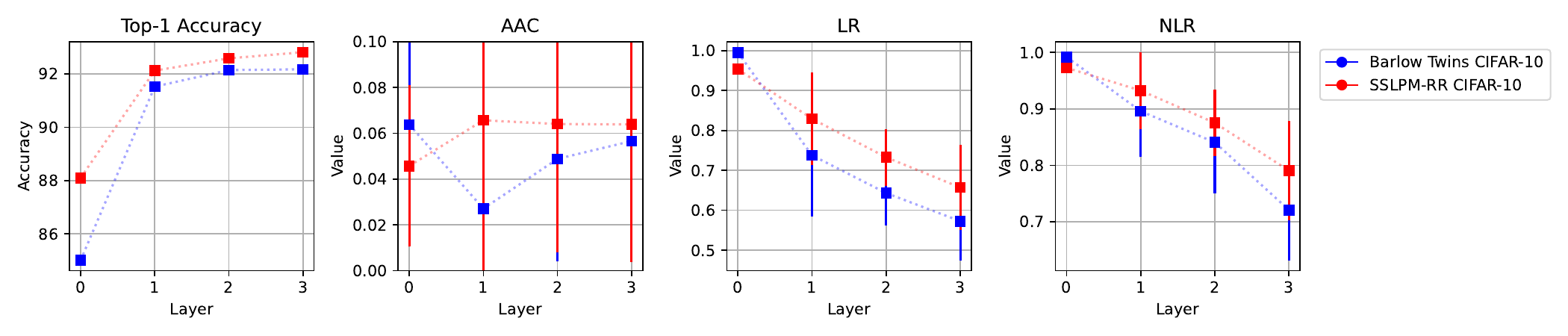}
  \caption{Ablation on the number of layers of the projector for CIFAR-10. Redundancy measures have the standard deviation over all measured indices shown.}
  \label{fig:layers_of_projector_cifar10}
\end{figure}

The projector, which is an MLP that sits between the embedding space and the space on which the loss is applied, as seen in Figure \ref{fig:main_model_diagram}, has been shown to be of crucial importance for many SSL methods \citep{gupta2022understanding} where no projector or a less expressive projector usually results in lower accuracy. 

In Figure \ref{fig:layers_of_projector_cifar10} we have measured the three redundancy values for Barlow Twins and SSLPM-RR on CIFAR-10 for different projector depths. Interestingly, we find that SSLPM-RR accuracy, compared to Barlow Twins, is less affected by the number of layers in the projectors. 
Further, we find it noteworthy that although AAC reveals no clear trends regarding redundancy and accuracy, we see a very clear trend for LR and NLR. As the number of projector layers is increased, the amount of redundancy removed from the embeddings increases. This extends work by \citep{gupta2022understanding} and provides a novel perspective on the effects of projectors.

Figure \ref{fig:layers_of_projector_cifar100_imagenet100} in Appendix \ref{app:further_analysis} show analogous results on CIFAR-100 and ImageNet-100.

\subsection{Redundancy in other SSL models}

\begin{figure}[h]
  \centering
  \includegraphics[width=\textwidth]{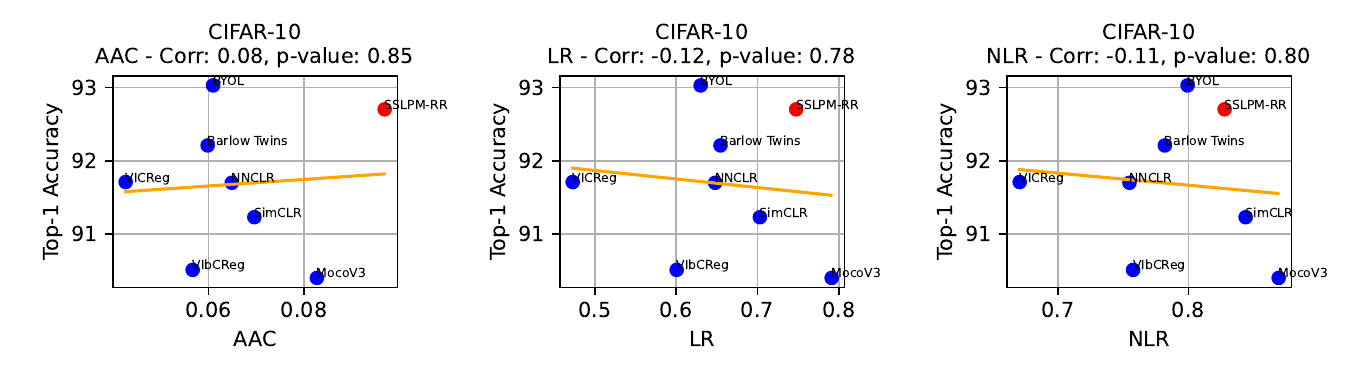}
  \caption{AAC, LR, and NLR embedding space redundancies plotted against Top-1 accuracy for different SSL methods on CIFAR-10. For the best fit lines, Pearson correlation and $p$-value for the null hypothesis that the data is uncorrelated, is reported.}
  \label{fig:redundancy_cifar10}
\end{figure}

Given the strong relationship between redundancy and performance for Barlow Twins and SSLPM, one might be tempted to think that low redundancy is a feature of \emph{all} high quality embeddings, whatever the pretraining approach.
The evidence presented in Figure \ref{fig:redundancy_cifar10} for CIFAR-10 and Figure \ref{fig:redundancy_cifar100_imagenet_100} in \cref{sec:additional_evidence} for CIFAR-100 does not support this conclusion. 
P values of $0.79$ and $0.93$ indicate that likely no linear relationship is present (looking at LR vs Top-1 accuracy). Further, the results are inconsistent across the redundancy measures. 

However, in the case of ImageNet-100, also in Figure \ref{fig:redundancy_cifar100_imagenet_100}, the data tells a different story with negative correlations between the Top-1 accuracy and all redundancy measures and low p values (p < 0.05 for LR and NLR). 

From the provided evidence, we cannot conclusively say that embeddings from other training approaches will also exhibit a negative correlation between redundancy and performance.

Nevertheless, across all datasets, we consistently find that all methods outperforming SSLPM-RR, namely BYOL, Barlow Twins, VICReg, and NNCLR, exhibit significantly less redundancy in their embeddings. Given that these methods do not perform active redundancy reduction, we see this as evidence that high performing models perform implicit redundancy reduction.

\section{Conclusion}

Although the idea of embedding space redundancies goes back to \citet{schmidhuber1996semilinear}, contemporary work on SSL only considers a narrow notion of redundancy.
With our contribution, we aim to close this gap in the SSL literature by introducing a framework for quantifying and reducing redundancy in embedding spaces. 

We introduce a hierarchy of redundancy measures and establish their theoretical foundations. 
% for these measures are related and have been able to show that high performing methods need not feature minimal redundancy.
Furthermore, we conduct an extensive empirical study to investigate the relationship between downstream model performance and embedding space redundancy. We find evidence for an inverse relationship between redundancy and performance when removing linear redundancy.  
With the introduction of SSLPM, we demonstrate the feasibility of reducing higher-order redundancy relations.
% than are more complex than simple pairwise correlations, as done in Barlow Twins. 
However, we find that there is currently no evidence that reducing more complex redundancies is beneficial for downstream model performance.

\section{Limitations and Future Work}
\label{sec:future_work}

% Although the idea of embedding space redundancies goes back to \citet{schmidhuber1996semilinear}, there is little contemporary work investigating the role and impact of redundancy. 
% As SSL methods have become central, e.g. for transformer-based methods \citep{vaswani2017attention}, we expect this discussion to become increasingly relevant.

Despite pairwise correlations (AAC) only offering a narrow view of redundancy in embedding spaces, we have not found evidence that reducing more general redundancies, such as LR (SSLPM-RR) or NLR (SSLPM-SGD), results in better performance. Our empirical results suggest that there is a limit to the redundancy that can be extracted as when using a large weight (large $\lambda$) for the redundancy term in the loss function, both Barlow Twins and SSLPM suffer from model collapse leading to degrading performance.
% Where these limits lie is unclear to us and warrants follow-up work.

Future work could explore whether similar results can be found for other data modalities, such as audio, text or multi-modal SSL.

\bibliography{main}

\begin{thebibliography}{26}
\providecommand{\natexlab}[1]{#1}
\providecommand{\url}[1]{\texttt{#1}}
\expandafter\ifx\csname urlstyle\endcsname\relax
  \providecommand{\doi}[1]{doi: #1}\else
  \providecommand{\doi}{doi: \begingroup \urlstyle{rm}\Url}\fi

\bibitem[Assran et~al.(2023)Assran, Duval, Misra, Bojanowski, Vincent, Rabbat, LeCun, and Ballas]{assran2023self}
Mahmoud Assran, Quentin Duval, Ishan Misra, Piotr Bojanowski, Pascal Vincent, Michael Rabbat, Yann LeCun, and Nicolas Ballas.
\newblock Self-supervised learning from images with a joint-embedding predictive architecture.
\newblock In \emph{Proceedings of the IEEE/CVF Conference on Computer Vision and Pattern Recognition}, pp.\  15619--15629, 2023.

\bibitem[Bardes et~al.(2021)Bardes, Ponce, and LeCun]{bardes2021vicreg}
Adrien Bardes, Jean Ponce, and Yann LeCun.
\newblock Vicreg: Variance-invariance-covariance regularization for self-supervised learning.
\newblock \emph{arXiv preprint arXiv:2105.04906}, 2021.

\bibitem[Chen et~al.(2020)Chen, Kornblith, Norouzi, and Hinton]{pmlr-v119-chen20j}
Ting Chen, Simon Kornblith, Mohammad Norouzi, and Geoffrey Hinton.
\newblock A simple framework for contrastive learning of visual representations.
\newblock In Hal~Daumé III and Aarti Singh (eds.), \emph{Proceedings of the 37th International Conference on Machine Learning}, volume 119 of \emph{Proceedings of Machine Learning Research}, pp.\  1597--1607. PMLR, 13--18 Jul 2020.
\newblock URL \url{https://proceedings.mlr.press/v119/chen20j.html}.

\bibitem[Chen \& He(2020)Chen and He]{chen2020exploring}
Xinlei Chen and Kaiming He.
\newblock Exploring simple siamese representation learning. in 2021 ieee.
\newblock In \emph{CVF conference on computer vision and pattern recognition (CVPR)}, pp.\  15745--15753, 2020.

\bibitem[Chen et~al.(2021)Chen, Xie, and He]{chen2021mocov3}
Xinlei Chen, Saining Xie, and Kaiming He.
\newblock An empirical study of training self-supervised vision transformers.
\newblock In \emph{Proceedings of the IEEE/CVF international conference on computer vision}, pp.\  9640--9649, 2021.

\bibitem[Da~Costa et~al.(2022)Da~Costa, Fini, Nabi, Sebe, and Ricci]{da2022solo}
Victor G~Turrisi Da~Costa, Enrico Fini, Moin Nabi, Nicu Sebe, and Elisa Ricci.
\newblock Solo-learn: A library of self-supervised methods for visual representation learning.
\newblock \emph{The Journal of Machine Learning Research}, 23\penalty0 (1):\penalty0 2521--2526, 2022.

\bibitem[Deng et~al.(2009)Deng, Dong, Socher, Li, Li, and Fei-Fei]{deng2009imagenet}
Jia Deng, Wei Dong, Richard Socher, Li-Jia Li, Kai Li, and Li~Fei-Fei.
\newblock Imagenet: A large-scale hierarchical image database.
\newblock In \emph{2009 IEEE conference on computer vision and pattern recognition}, pp.\  248--255. Ieee, 2009.

\bibitem[Devlin et~al.(2018)Devlin, Chang, Lee, and Toutanova]{devlin2018bert}
Jacob Devlin, Ming-Wei Chang, Kenton Lee, and Kristina Toutanova.
\newblock Bert: Pre-training of deep bidirectional transformers for language understanding.
\newblock \emph{arXiv preprint arXiv:1810.04805}, 2018.

\bibitem[Dwibedi et~al.(2021)Dwibedi, Aytar, Tompson, Sermanet, and Zisserman]{dwibedi2021little}
Debidatta Dwibedi, Yusuf Aytar, Jonathan Tompson, Pierre Sermanet, and Andrew Zisserman.
\newblock With a little help from my friends: Nearest-neighbor contrastive learning of visual representations.
\newblock In \emph{Proceedings of the IEEE/CVF International Conference on Computer Vision}, pp.\  9588--9597, 2021.

\bibitem[Ermolov et~al.(2021)Ermolov, Siarohin, Sangineto, and Sebe]{ermolov2021whitening}
Aleksandr Ermolov, Aliaksandr Siarohin, Enver Sangineto, and Nicu Sebe.
\newblock Whitening for self-supervised representation learning.
\newblock In \emph{International Conference on Machine Learning}, pp.\  3015--3024. PMLR, 2021.

\bibitem[Goodfellow et~al.(2014)Goodfellow, Pouget-Abadie, Mirza, Xu, Warde-Farley, Ozair, Courville, and Bengio]{goodfellow2014generative}
Ian Goodfellow, Jean Pouget-Abadie, Mehdi Mirza, Bing Xu, David Warde-Farley, Sherjil Ozair, Aaron Courville, and Yoshua Bengio.
\newblock Generative adversarial nets.
\newblock \emph{Advances in neural information processing systems}, 27, 2014.

\bibitem[Grill et~al.(2020)Grill, Strub, Altch{\'e}, Tallec, Richemond, Buchatskaya, Doersch, Avila~Pires, Guo, Gheshlaghi~Azar, et~al.]{grill2020bootstrap}
Jean-Bastien Grill, Florian Strub, Florent Altch{\'e}, Corentin Tallec, Pierre Richemond, Elena Buchatskaya, Carl Doersch, Bernardo Avila~Pires, Zhaohan Guo, Mohammad Gheshlaghi~Azar, et~al.
\newblock Bootstrap your own latent-a new approach to self-supervised learning.
\newblock \emph{Advances in neural information processing systems}, 33:\penalty0 21271--21284, 2020.

\bibitem[Gupta et~al.(2022)Gupta, Ajanthan, Hengel, and Gould]{gupta2022understanding}
Kartik Gupta, Thalaiyasingam Ajanthan, Anton van~den Hengel, and Stephen Gould.
\newblock Understanding and improving the role of projection head in self-supervised learning.
\newblock \emph{arXiv preprint arXiv:2212.11491}, 2022.

\bibitem[He et~al.(2016)He, Zhang, Ren, and Sun]{he2016deep}
Kaiming He, Xiangyu Zhang, Shaoqing Ren, and Jian Sun.
\newblock Deep residual learning for image recognition.
\newblock In \emph{Proceedings of the IEEE conference on computer vision and pattern recognition}, pp.\  770--778, 2016.

\bibitem[He et~al.(2020)He, Fan, Wu, Xie, and Girshick]{he2020momentum}
Kaiming He, Haoqi Fan, Yuxin Wu, Saining Xie, and Ross Girshick.
\newblock Momentum contrast for unsupervised visual representation learning.
\newblock In \emph{Proceedings of the IEEE/CVF conference on computer vision and pattern recognition}, pp.\  9729--9738, 2020.

\bibitem[Hsu et~al.(2021)Hsu, Bolte, Tsai, Lakhotia, Salakhutdinov, and Mohamed]{hsu2021hubert}
Wei-Ning Hsu, Benjamin Bolte, Yao-Hung~Hubert Tsai, Kushal Lakhotia, Ruslan Salakhutdinov, and Abdelrahman Mohamed.
\newblock Hubert: Self-supervised speech representation learning by masked prediction of hidden units.
\newblock \emph{IEEE/ACM Transactions on Audio, Speech, and Language Processing}, 29:\penalty0 3451--3460, 2021.

\bibitem[Lee \& Aune(2021)Lee and Aune]{lee2021computer}
Daesoo Lee and Erlend Aune.
\newblock Computer vision self-supervised learning methods on time series.
\newblock \emph{arXiv preprint arXiv:2109.00783}, 2021.

\bibitem[Loshchilov \& Hutter(2019)Loshchilov and Hutter]{loshchilov2018fixing}
Ilya Loshchilov and Frank Hutter.
\newblock Decoupled weight decay regularization.
\newblock In \emph{International Conference on Learning Representations}, 2019.
\newblock URL \url{https://openreview.net/forum?id=Bkg6RiCqY7}.

\bibitem[Oord et~al.(2018)Oord, Li, and Vinyals]{oord2018representation}
Aaron van~den Oord, Yazhe Li, and Oriol Vinyals.
\newblock Representation learning with contrastive predictive coding.
\newblock \emph{arXiv preprint arXiv:1807.03748}, 2018.

\bibitem[Schmidhuber(2020)]{schmidhuber2020generative}
J{\"u}rgen Schmidhuber.
\newblock Generative adversarial networks are special cases of artificial curiosity (1990) and also closely related to predictability minimization (1991).
\newblock \emph{Neural Networks}, 127:\penalty0 58--66, 2020.

\bibitem[Schmidhuber et~al.(1996)Schmidhuber, Eldracher, and Foltin]{schmidhuber1996semilinear}
J{\"u}rgen Schmidhuber, Martin Eldracher, and Bernhard Foltin.
\newblock Semilinear predictability minimization produces well-known feature detectors.
\newblock \emph{Neural Computation}, 8\penalty0 (4):\penalty0 773--786, 1996.

\bibitem[Schraudolph et~al.(1999)Schraudolph, Eldracher, and Schmidhuber]{schraudolph1999processing}
Nicol~N Schraudolph, Martin Eldracher, and J{\"u}rgen Schmidhuber.
\newblock Processing images by semi-linear predictability minimization.
\newblock \emph{Network: Computation in Neural Systems}, 10\penalty0 (2):\penalty0 133, 1999.

\bibitem[Shwartz~Ziv \& LeCun(2024)Shwartz~Ziv and LeCun]{e26030252}
Ravid Shwartz~Ziv and Yann LeCun.
\newblock To compress or not to compress—self-supervised learning and information theory: A review.
\newblock \emph{Entropy}, 26\penalty0 (3), 2024.
\newblock ISSN 1099-4300.
\newblock \doi{10.3390/e26030252}.
\newblock URL \url{https://www.mdpi.com/1099-4300/26/3/252}.

\bibitem[Tsai et~al.(2021)Tsai, Bai, Morency, and Salakhutdinov]{tsai2021note}
Yao-Hung~Hubert Tsai, Shaojie Bai, Louis-Philippe Morency, and Ruslan Salakhutdinov.
\newblock A note on connecting barlow twins with negative-sample-free contrastive learning.
\newblock \emph{arXiv preprint arXiv:2104.13712}, 2021.

\bibitem[You et~al.(2017)You, Gitman, and Ginsburg]{you2017large}
Yang You, Igor Gitman, and Boris Ginsburg.
\newblock Large batch training of convolutional networks.
\newblock \emph{arXiv preprint arXiv:1708.03888}, 2017.

\bibitem[Zbontar et~al.(2021)Zbontar, Jing, Misra, LeCun, and Deny]{zbontar2021barlow}
Jure Zbontar, Li~Jing, Ishan Misra, Yann LeCun, and St{\'e}phane Deny.
\newblock Barlow twins: Self-supervised learning via redundancy reduction.
\newblock In \emph{International Conference on Machine Learning}, pp.\  12310--12320. PMLR, 2021.

\end{thebibliography}
\bibliographystyle{tmlr}

\appendix

\section{Details on Measuring Redundancy}
\label{app:redundancy_measures}

For all trained models, we have extracted the embeddings of all test and train samples. The two datasets were then individually standardized such that every embedding feature has a mean of 0 and a standard deviation of 1. 

Hence, let $\vec{X}\in\R^{k}$ be the output of an encoder, for example the $k=512$ dimensional output of a ResNet-18, right before the last fully-connected layer. For practical purposes, we always standardize a batch of $n$ embeddings $\mathbf{X}\in \R^{n\times k}$ such that for all dimensions $i$ we have $\E[(\textnormal{X})_i] = 0$ and $\Var[(\textnormal{X})_i] = 1$. Given the standardized embeddings, we define the empirical covariance matrix $\mathbf{C}:=  \frac{\mathbf{X}^T\cdot \mathbf{X}}{n}$.

For AAC, the values from the empirical covariance matrix of the standardized embeddings are used.
For LR and NLR a random sample of 20\% of all neurons/features is taken (20\% of 512 $\approx$ 102). Then using 80\% of the data, a model was trained to predict the neuron's activation and then evaluated on the remaining 20\%.

For NLR, we used a three layer MLP with a fixed number of training steps and for LR ridge regression where we optimized over the ridge penalty.

In the figures, we report the mean of the measures over all neurons predicted.

The redundancy measurement experiments used a NVIDIA RTX 3090 card and took between 30 minutes and 2 hours per dataset and method.

\section{Modification of SSLPM-RR Objective}
\label{app:SSLPMREG}

In our main analysis for SSLPM-RR we have defined $\bm{W}$ as
\begin{align}
\bm{W}_{\text{default}}  &= \argmin_{\bm{W}} \norm{R_{\bm{W}}(\mathbf{\Bar{M}} \odot {\mathbf{X_{A||B}}}) - {\mathbf{X_{A||B}}}}_F^2 + \mu \norm{{\bm{W}}}_F^2
\end{align}
with
\begin{align}
    \bm{W}_{\text{default}} &= \left(\left(\left(\mathbf{\Bar{M}} \odot \mathbf{X_{A||B}}\right)^T \cdot \left(\mathbf{\Bar{M}} \odot \mathbf{X_{A||B}}\right) + \mu \mI \right)^{-1} \cdot \left(\mathbf{\Bar{M}} \odot \mathbf{X_{A||B}}\right)^T \cdot \mathbf{X_{A||B}} \right)
\end{align}
which enforces $\bm{W}$ to reconstruct the complete representations $\mathbf{X_{A||B}}$ from the masked representations. In particular, this means, that non-masked entries should be kept as is (i.e. ideally an identity function applied to them).

We can change this and require that we only reconstruct the masked representations and map all non-masked representations to zero, that way we no longer incentivize learning an identity mapping. 

This results in the following optimization problem for $\bm{W}$:
\begin{align}
\bm{W}_{\text{0}}  &= \argmin_{\bm{W}} \norm{R_{\bm{W}}(\mathbf{\Bar{M}} \odot {\mathbf{X_{A||B}}}) - \mathbf{M} \odot {\mathbf{X_{A||B}}}}_F^2 + \mu \norm{{\bm{W}}}_F^2\end{align}
with
\begin{align}
    \bm{W}_{\text{0}} &= \left(\left(\left(\mathbf{\Bar{M}} \odot \mathbf{X_{A||B}}\right)^T \cdot \left(\mathbf{\Bar{M}} \odot \mathbf{X_{A||B}}\right) +  \mu \mI \right)^{-1} \cdot \left(\mathbf{\Bar{M}} \odot \mathbf{X_{A||B}}\right)^T \cdot \left(\mathbf{M} \odot \mathbf{X_{A||B}}\right) \right)
\end{align}
During training on CIFAR-10 we barely see a difference between the two models as shown in Table \ref{tab:masking_reg_obj_change}. They feature a nearly identical final predictability loss as well as accuracies on CIFAR-10. Hence, we conclude that the exact choice of the objective for the ridge regression is likely not of crucial importance for our method.

\begin{table}[h]
\centering

\caption{Change of regression objective}
\label{tab:masking_reg_obj_change}

\begin{tabular}{@{}lccc@{}}
\toprule
             & \multicolumn{3}{c}{CIFAR-10} \\
\cmidrule(lr){2-4} 
Models       & Top 1         & Top 5        & Last Epoch Avg $\mathcal{L}_{pred}$  \\
\midrule
SSLPM-RR with $\bm{W}_{\text{default}}$  &  92.81  &    99.87         &    0.65     \\
SSLPM-RR with $\bm{W}_{\text{0}}$     &   92.75    &    99.82   & 0.66  \\
\bottomrule
\end{tabular}
\end{table}

\section{Further Results \& Analysis}\label{app:further_analysis}

\subsection{Online vs Offline Evaluation}
\label{app:offline_vs_online}
Following the methodology by \cite{da2022solo} we report online as well as offline results for ImageNet-100.
While in the online case a linear prediction head is also trained during pre-training, bur without gradient flowing through the prediction head into the encoder, the off-line linear prediction head is trained for 100 epochs on the training data after the backbone has finished (pre-)training and its weights have been completely frozen. Unexpectedly, the differences between the two evaluation modes is within the margin of error. 

\begin{table}[h]
\setlength{\tabcolsep}{5pt}
\centering
\caption{Evaluation of different models on various datasets. \textbf{\underline{Top1}}, \textbf{Top2}, and \underline{Top3} models are highlighted per dataset.}
\label{table:resModelsOnlineOffline}
\footnotesize
\begin{tabular}{@{}lcc|cc@{}}
\toprule
             & \multicolumn{2}{c}{ImageNet-100 {(Online)}} &\multicolumn{2}{c}{ImageNet-100 {(Offline)}} \\
\cmidrule(lr){2-3} \cmidrule(lr){4-5} 
Models            & Top 1           & Top 5     & Top 1    & Top 5    \\
\midrule
Barlow Twins & \textbf{\underline{80.31 (0.099)}} & \textbf{\underline{95.33 (0.147)}}  &  \textbf{\underline{80.24 (0.179)}}  &  \textbf{\underline{95.21 (0.058)}}     \\
BYOL         &  76.12 &         93.92  &   76.66   &  93.74   \\
NNCLR        &  \underline{79.22} & 94.82 &  \underline{79.58}  &  \underline{94.9} \\
SimCLR       &  77.38 & 94.02 &  77.50  &  93.86   \\
MocoV3       &  74.72 & 93.02 &  74.46   &  92.86    \\
VICReg        & \textbf{79.56} & \textbf{95.10} & 79.28 & 94.52  \\
VIbCReg       & 78.14 & 94.02 & 78.26   & 93.96 \\
\textbf{SSLPM-RR (ours)} & 79.17 (0.310)    &   \underline{95.02 (0.178)}  & \textbf{79.99 (0.711)}  &  \textbf{94.95 (0.253)}  \\
\textbf{SSLPM-SGD 1l} & 76.19 (0.196)	& 93.90 (0.193)	&76.73 (0.320)	&93.83 (0.147)   \\
\textbf{SSLPM-SGD 3l} &  64.27 (0.594)	&88.55 (0.600)	&67.93 (0.938)	&90.79 (0.659) \\
\bottomrule
\end{tabular}
\end{table}

\begin{table}[h]
\centering
\caption{Ablation showing the importance of the redundancy reduction term.}
\label{table:lambda_ablation_onlin_offline}
\footnotesize
\begin{tabular}{@{}lcc|cc@{}}
\toprule
             & \multicolumn{2}{c}{ImageNet-100 {(Online)}} &\multicolumn{2}{c}{ImageNet-100 {(Offline)}} \\
\cmidrule(lr){2-3} \cmidrule(lr){4-5} 
Models           & Top 1           & Top 5     & Top 1    & Top 5    \\
\midrule
{SSLPM-RR} &  79.17 (0.310)    &   95.02 (0.178)  & 79.99 (0.711)  &   94.95 (0.253)  \\
{SSLPM-SGD 1l (ours)}  & 76.19 (0.196)	& 93.90 (0.193)	&76.73 (0.320)	&93.83 (0.147)   \\
{SSLPM-SGD 3l (ours)} & 64.27 (0.594)	&88.55 (0.600)	& 67.93 (0.938)	&90.79 (0.659) \\
SSLPM ($\lambda = 0$) &    50.81 (0.388)      &   79.86 (0.046)   &  57.78 (0.597)  & 84.23 (0.423)    \\
\bottomrule
\end{tabular}
\end{table}

\subsection{Training Behavior with Different Loss Weights in SSLPM}
\label{app:impact_of_lamba_on_training}
\begin{figure}[h]
    \centering
    \begin{minipage}{0.24\textwidth}
        \centering
        \includegraphics[width=\textwidth]{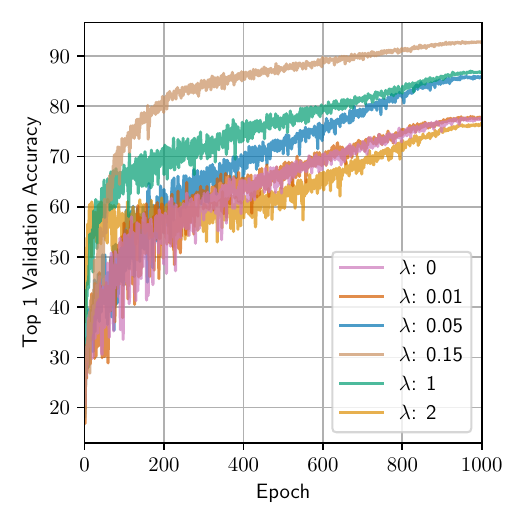}
    \end{minipage}\hfill
    \begin{minipage}{0.24\textwidth}
        \centering
        \includegraphics[width=\textwidth]{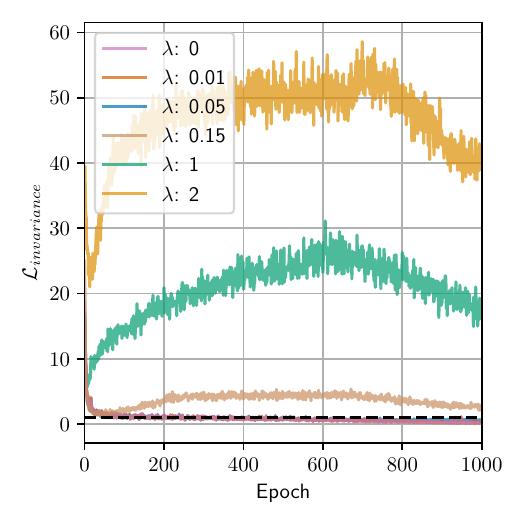}
    \end{minipage}\hfill
    \begin{minipage}{0.24\textwidth}
        \centering
        \includegraphics[width=\textwidth]{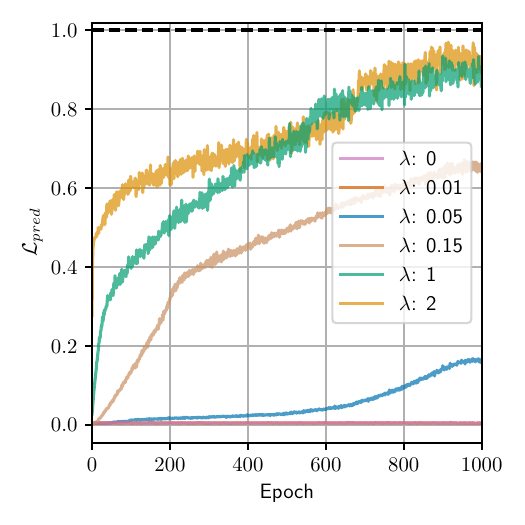}
    \end{minipage}\hfill
    \begin{minipage}{0.24\textwidth}
        \centering
        \includegraphics[width=\textwidth]{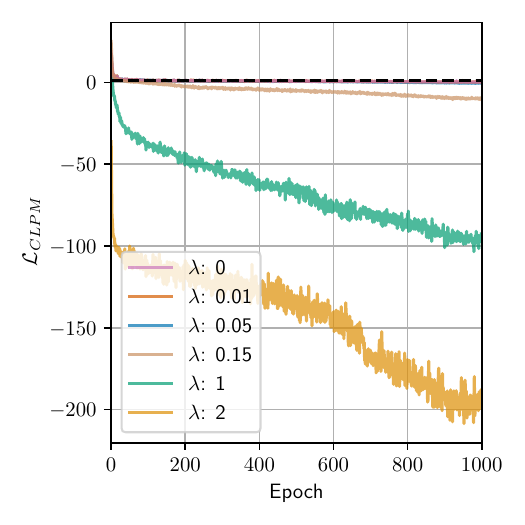}
    \end{minipage}
        \caption{Training dynamics for different $\lambda$: (1) accuracy. (2) $\mathcal{L}_\text{invariance}$ (3) $\mathcal{L}_\text{pred}$. (4) $\mathcal{L}_\text{SSLPM}$.}
    \label{fig:lambda_comparison}

\end{figure}

The hyperparameter $\lambda$ as introduced in Equation \ref{eq:clLoss} plays a crucial role in weighing the different losses against one another. In Figure \ref{fig:lambda_comparison} we plot different loss metrics over training for different $\lambda$'s. We see that a $\lambda$ that is too small (such as 0.01) performs similar to the case where we have $\lambda = 0$. For $\lambda = 1, 2$ we see that $\mathcal{L}_\text{pred}$ is close to 1, but this does not lead to the best observed performance. Interestingly, we find that over the course of training even though $\mathcal{L}_\text{SSLPM}$ decreases steadily, first $\mathcal{L}_\text{invariance}$ nearly vanishes, but then increases again as $\mathcal{L}_\text{pred}$ is optimized. This dynamic is especially present for large $\lambda$.

The observations from Figure \ref{fig:lambda_comparison} allow us to conclude that there seems to be a tradeoff between $\mathcal{L}_\text{invariance}$ and $\mathcal{L}_\text{pred}$ whereby too much redundancy reduction hurts model performance.

\subsection{Effect of Ridge Penalty and Batch Size in SSLPM-RR}

\begin{figure}
  \centering
  \includegraphics[width=\textwidth]{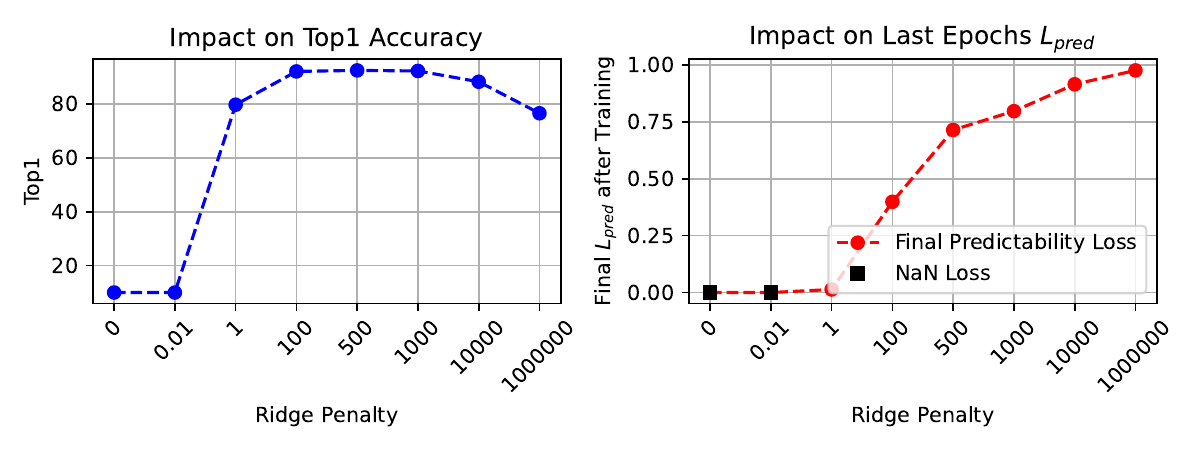}
  \caption{Impact of ridge penalty on Top-1 CIFAR-10 accuracy.}
  \label{fig:ridge_ablation}
\end{figure}

When the batch size smaller than half the size of the representation dimension on which we calculate the loss, standard linear regression can no longer be performed as the system is singular. To circumvent this and to increase regression stability, we introduce a ridge penalty. The experiment visualized in Figure \ref{fig:ridge_ablation} uses a constant batch size of 64 and increasing ridge penalties. We see that a nonexistent or too small ridge penalty results in model collapse due to the regression being singular. However, an appropriately chosen ridge penalty allows the model to learn as if it was operating in a non-singular setting. 

Additionally, we show in Table \ref{tab:batch_size_top1} that SSLPM-RR performs well regardless of the batch size as long as an appropriate ridge penalty is chosen. 

\begin{table}[h]
\centering
\caption{Top-1 accuracy for different batch sizes on CIFAR-10.}
\label{tab:batch_size_top1}
\begin{tabular}{@{}lcc@{}}
\toprule
Batch Size & Top 1 Accuracy \\
\midrule
32 & 91.67 \\
64 & 92.51 \\
128 & 92.96 \\
256 & 92.76 \\
\bottomrule
\end{tabular}
\end{table}

\subsection{Impact of Masking Fraction on SSLPM-RR}

\begin{figure}[h]
    \centering
    \begin{minipage}{0.49\textwidth}
        \centering
        \includegraphics[width=\textwidth]{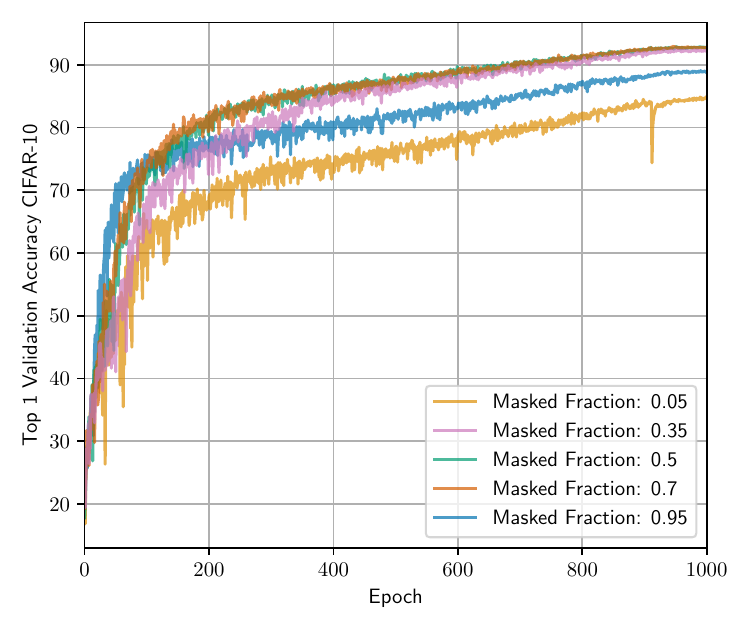}
    \end{minipage}\hfill
    \begin{minipage}{0.49\textwidth}
        \centering
        \includegraphics[width=\textwidth]{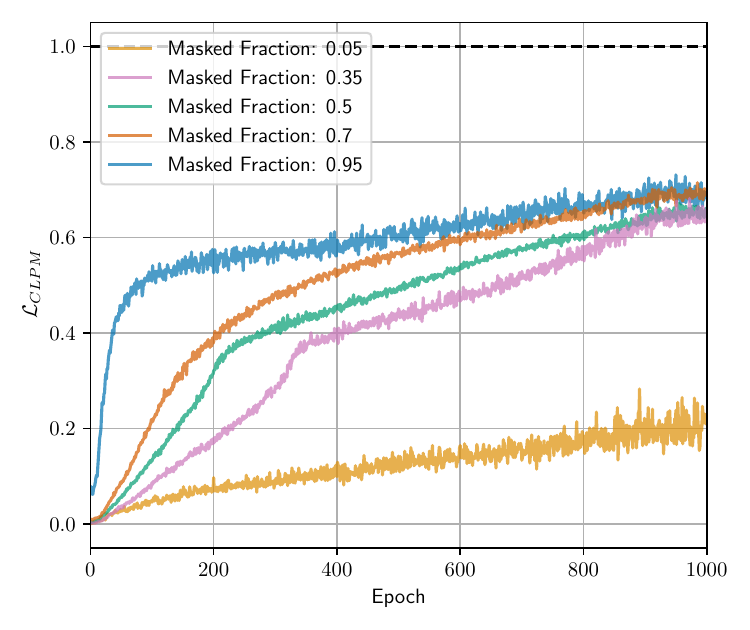}
    \end{minipage}
        \caption{(left) Accuracy over Training. (right) $\mathcal{L}_\text{pred}$ over Training}
    \label{fig:masking_fraction_comparison}
\end{figure}

The masking fraction, the percentage of representation's features to mask during training, is a key hyperparameter in SSLPM. In Figure \ref{fig:masking_fraction_comparison} we have trained multiple different models on CIFAR-10 while varying the masking fraction. We see that as long as the masking fraction is neither too small nor too large, training outcomes are not negatively impacted. Interestingly, masking 5\% performed significantly worse than masking 95\% of the features.

\subsection{The Empirical Relationship Between the Redundancy Measures}
\label{app:redundancy_measures_related}
\begin{figure}[h]
  \centering
  \includegraphics[width=\textwidth]{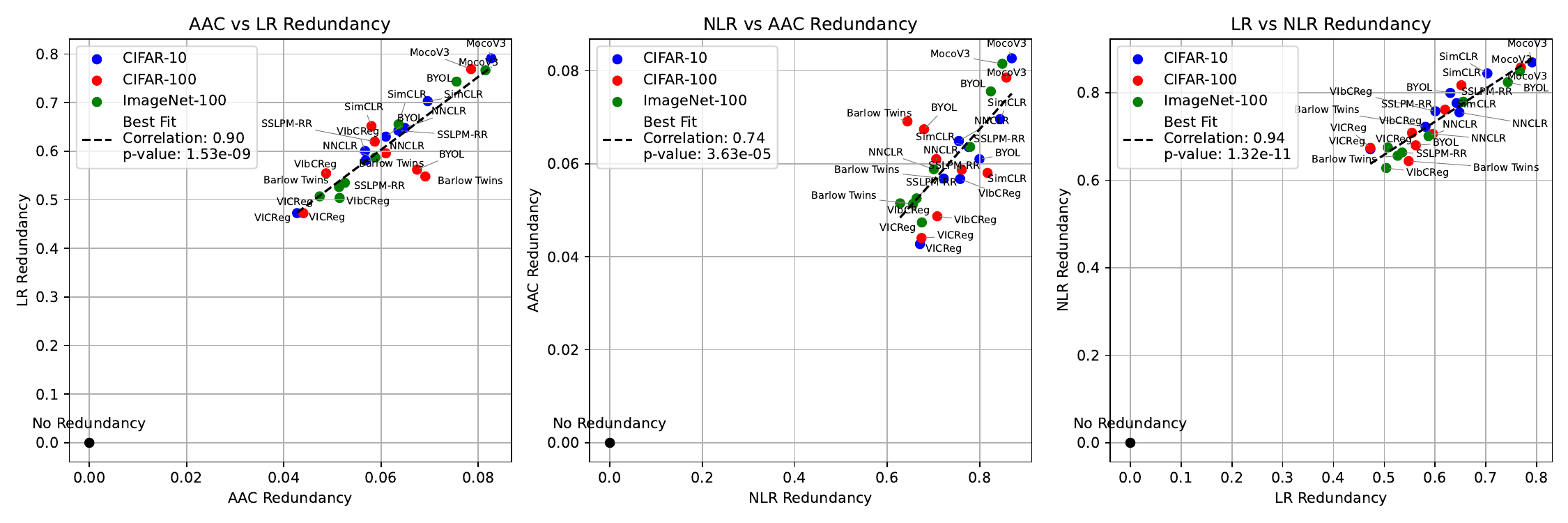}
  \caption{AAC, LR, and NLR plotted against each other for test set embeddings of different models on CIFAR-10, CIFAR-100, and ImageNet-100 datasets.}
  \label{fig:redundancies_different_datasets}
\end{figure}

Knowing how the redundancy measures are related in theory leaves the question of how the redundancy measures correlate in practice. To answer this question, we plot the redundancies, AAC, LR, and, NLR in pairwise fashion in Figure \ref{fig:redundancies_different_datasets} for all three datasets under investigation, CIFAR-10, CIFAR-100, and ImageNet-100.

From Figure \ref{fig:redundancies_different_datasets} we learn that the redundancy definitions correlate well, especially the more related ones. AAC and LR have a correlation coefficient of 0.90 whereas LR and NLR have a correlation coefficient of 0.94. 

Not surprisingly we find that for a given level of AAC redundancy we find higher levels of LR as well as for a given level of LR we find a higher level of NLR. This empirically validates our intuition with LR's ability to be high despite low AAC as well as NLR capturing nonlinear redundancies that LR cannot. 

% \subsection{Difference in Redundancies Between Test and Train Sets}
\subsection{Redundancies in the Test and Train Sets}

\begin{figure}[h]
  \centering
  \includegraphics[width=\textwidth]{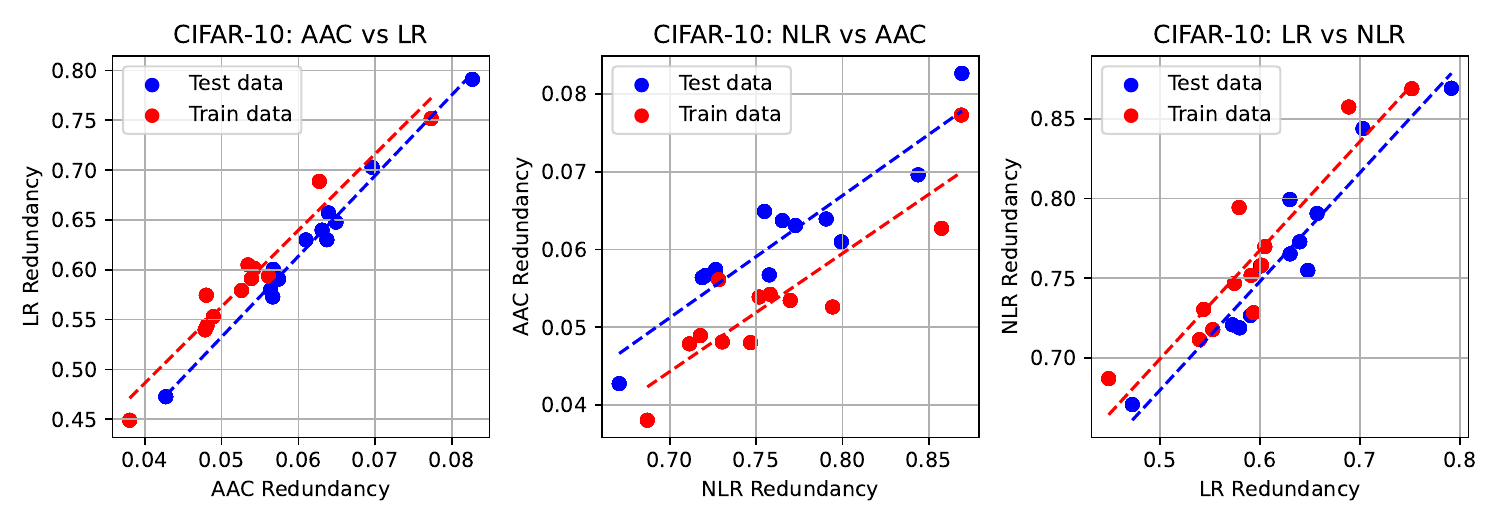}
  \caption{Difference in redundancy between test and train sets for CIFAR-10 embeddings.}
  \label{fig:test_train_cifar10}
\end{figure}

\begin{figure}[h]
  \centering
  \includegraphics[width=\textwidth]{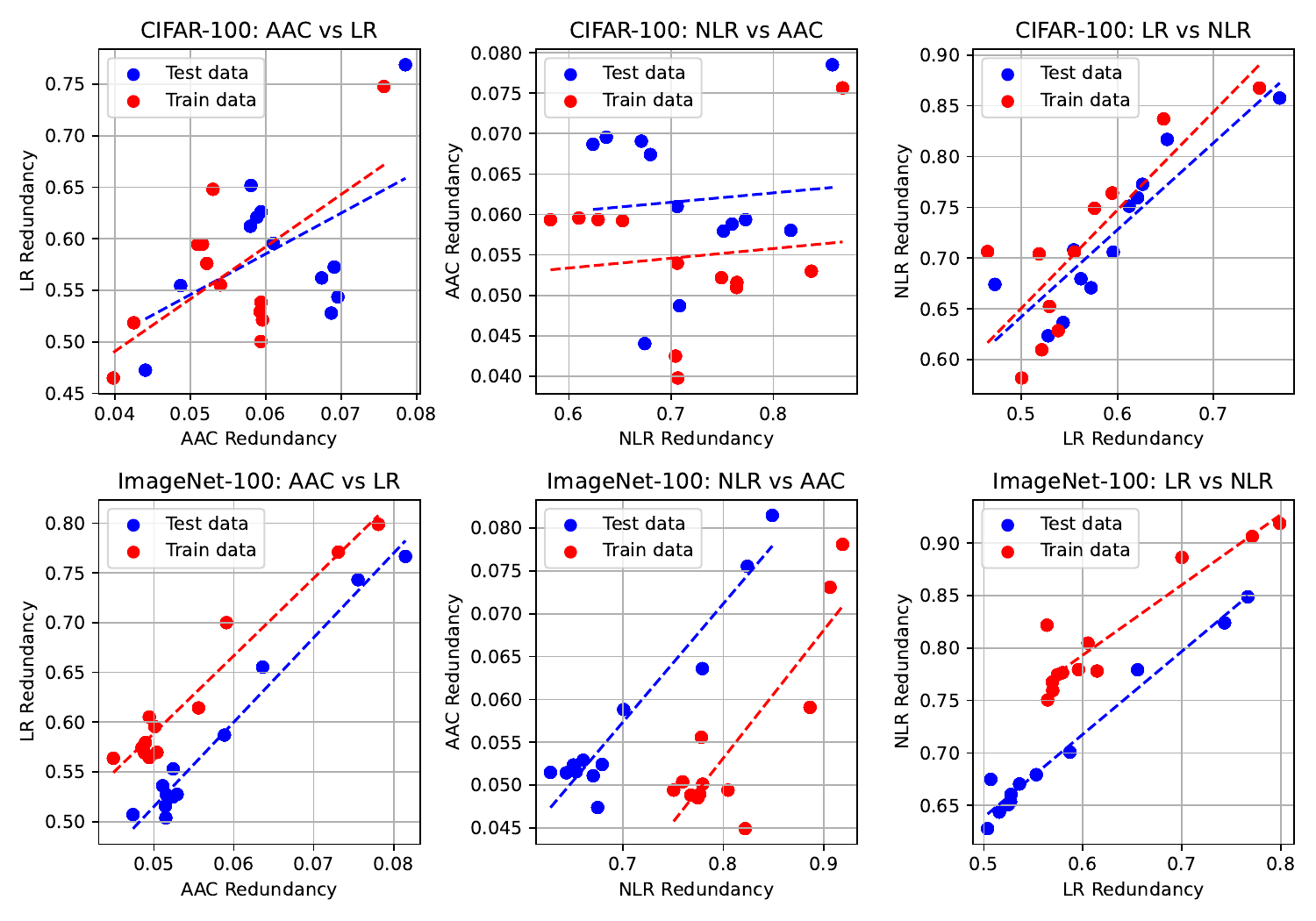}
  \caption{Difference in redundancy between test and train sets for CIFAR-100 and ImageNet-100 embeddings.}
  \label{fig:test_train_cifar100_imagenet100}
\end{figure}

So far, we have only looked at embeddings from test datasets. However, it begs the question if the results we see on the test sets also hold for the training dataset. To investigate this, we have collected the average redundancy values for all models under consideration and plotted them onto a pairwise redundancy chart. In Figure \ref{fig:test_train_cifar10} we see the CIFAR-10 train and test embeddings colored differently together with two best fit lines. We see that for a given level of AAC redundancy, there is more LR redundancy in the test than the train set. The same holds for NLR, where for a given level of LR redundancy we find more NLR redundancy in the test set. Figure \ref{fig:test_train_cifar100_imagenet100} shows that these findings transfer over to CIFAR-100 and ImageNet-100.

\section{Omitted Proofs}
\printProofs

\section{Further Implementation Details}
\label{app:implementation_details}

Certain parameters such as masking fraction, $\lambda$ and the ridge penalty were determined by a grid search over a logarithmic search space. Most of these results are included in the ablations. 

For optimization, the LARS optimizer is used \citep{you2017large} as well as a linear warm-up cosine annealing learning rate scheduler. Learning rates can be found in the accompanying code.

For all models except SSLPM-SGD a batch size of 256 images was used. In SSLPM-SGD we have used 512 to increase numerical stability. 

SSLPM-SGD is trained with the AdamW \citep{loshchilov2018fixing} optimizer using a base learning rate of $1e-3$ a weight decay of $1e-6$.

Further, different methods to enhance the predictor training stability were used:
\begin{itemize}
    \item The predictor does not have minibatches, we optimize with all data per batch.
    \item Instead of having the encoder optimize $ \mathcal{L}_{\text{invariance}} - \lambda \mathcal{L}_{\text{pred}}$ we optimize $ \mathcal{L}_{\text{invariance}} - \lambda  \log \left( \mathcal{L}_{\text{pred}}\right)$.
    \item For every optimization step of the encoder, the predictor can perform up to 500 optimization steps, but stopping early if MSE stops improving on the validation mask.
    \item We vary the learning rate of the predictor. For the first 25 batches, we increase the learning rate by multiplying with $1.005$. Afterward, we decrease it by dividing by 1.005.
    \item The predictability loss is clipped at 1 from above. 
\end{itemize}

A single training run of the models presented takes between 8 and 24 hours (depending on the model and  hyperparameters) on a modern GPU such as a NVIDIA RTX3090. Evaluating the redundancy takes around 1 hour per trained model. 

The ImageNet-100 dataset is the same as used in the solo-learn paper \citep{da2022solo} and is a 100 category subset of the ImageNet dataset \citet{deng2009imagenet}.

\section{Additional Evidence for Analysis \& Results}
\label{sec:additional_evidence}

\begin{figure}[h]
  \centering
  \includegraphics[width=\textwidth]{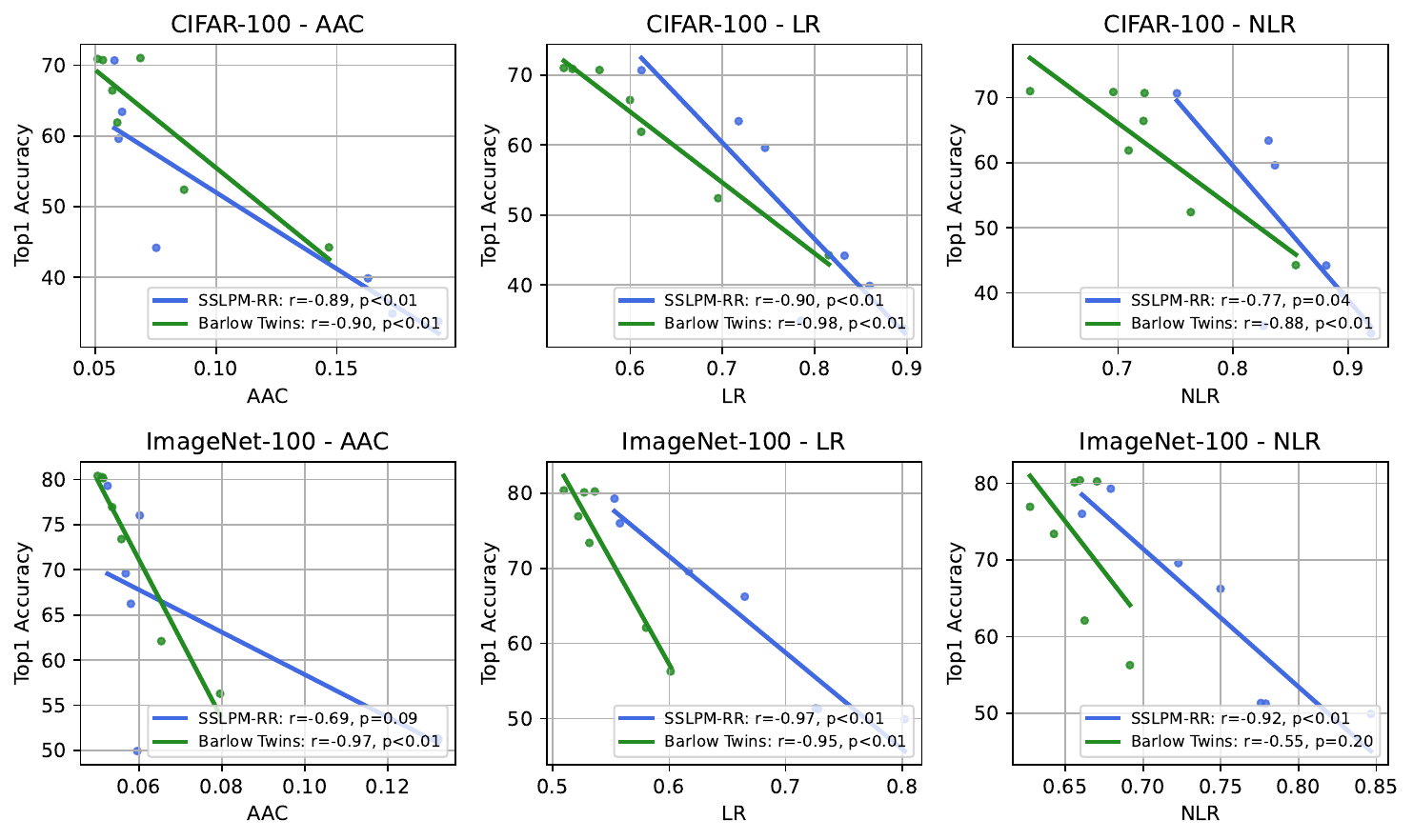}
  \caption{Relationship between Top-1 accuracy and different redundancy measures  redundancies on CIFAR-100 and ImageNet-100 plotted with the best fit line. For the best fit line Pearson correlation and $p$-value for the null hypothesis that the data is uncorrelated is reported.}
\label{fig:accuracy_redundancy_relationship_plot_cifar100_imagenet_100}
\end{figure}

\begin{figure}[h]
  \centering
  \includegraphics[width=\textwidth]{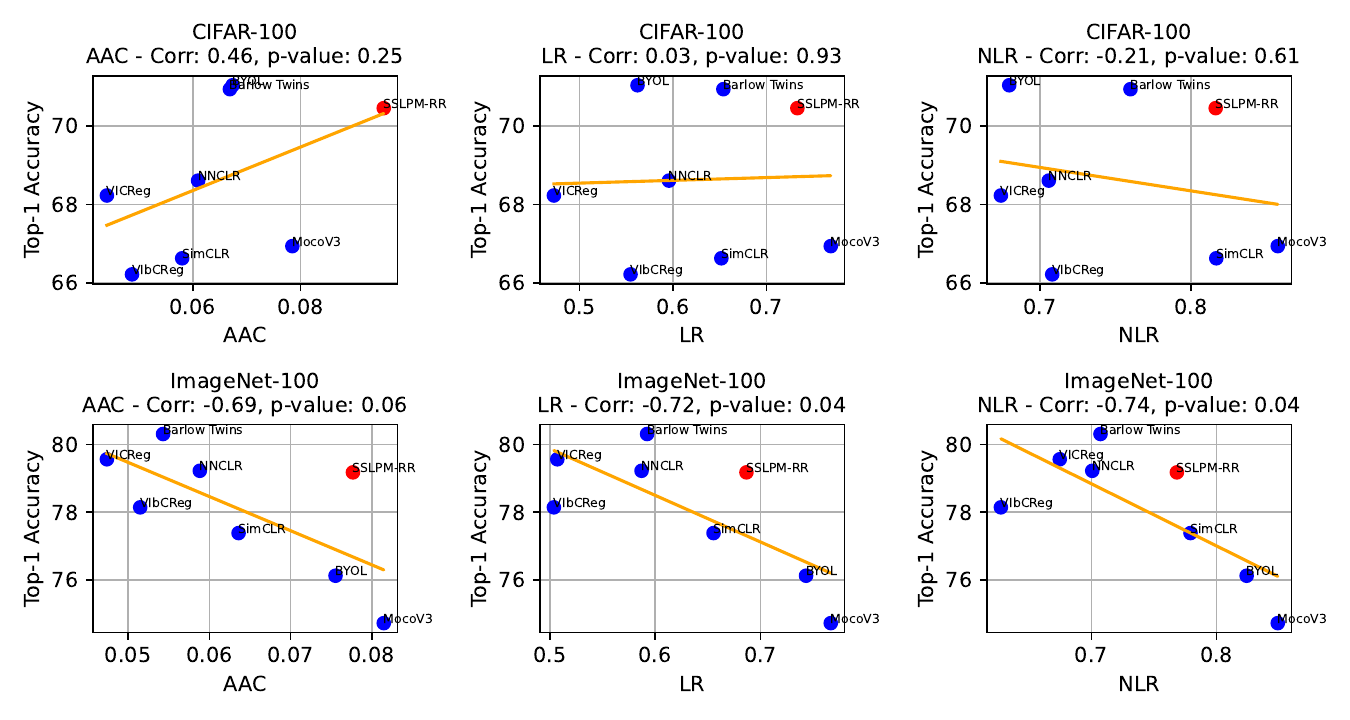}
  \caption{Redundancy measures vs. Top-1 accuracy of SSL methods on CIFAR-100, and ImageNet-100. For the best fit line Pearson correlation and $p$-value for the null hypothesis that the data is uncorrelated is reported.}
  \label{fig:redundancy_cifar100_imagenet_100}
\end{figure}

\begin{figure}[h]
  \centering
  \includegraphics[width=\textwidth]{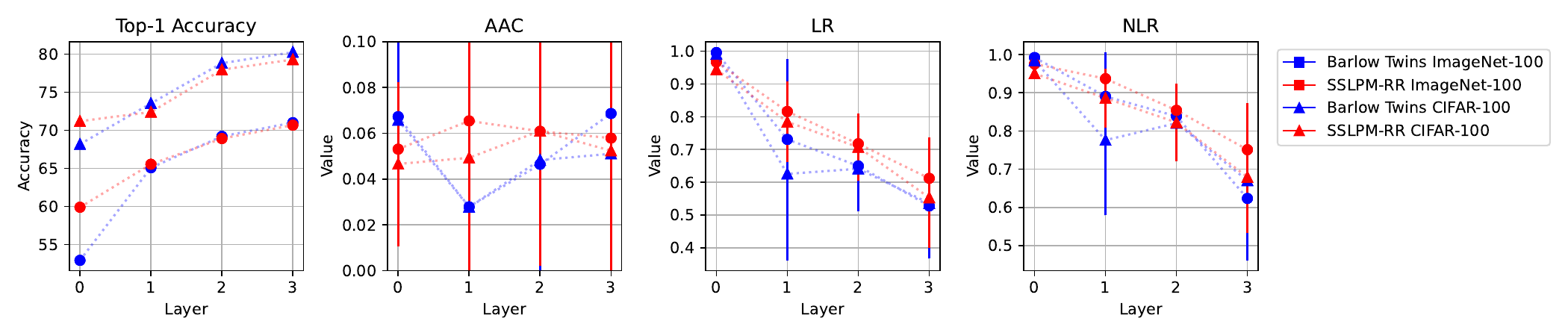}
  \caption{Ablation on the number of layers of the projector for CIFAR-100 and ImageNet-100. Redundancy measures have the standard deviation over all measured indices shown.}
  \label{fig:layers_of_projector_cifar100_imagenet100}
\end{figure}

\end{document}